%% file: main.tex
\documentclass[10pt,twocolumn,letterpaper]{article}

\usepackage[pagenumbers]{cvpr} % To force page numbers, e.g. for an arXiv version
\usepackage{footnote} % add this line to use the package

\usepackage{graphicx}
\usepackage{amsmath}
\usepackage{amssymb}
\usepackage{booktabs}
\usepackage{caption,subcaption}
\usepackage{soul}
\usepackage{xcolor}
\usepackage{pifont}
\usepackage{colortbl}
\usepackage[most]{tcolorbox}

\definecolor{Gray}{gray}{0.85}
\DeclareFontFamily{OT1}{pzc}{}
\DeclareFontShape{OT1}{pzc}{m}{it}{<-> s * [1.10] pzcmi7t}{}
\DeclareMathAlphabet{\mathpzc}{OT1}{pzc}{m}{it}
\usepackage[pagebackref,breaklinks,colorlinks]{hyperref}

\usepackage[capitalize]{cleveref}
\crefname{section}{Sec.}{Secs.}
\Crefname{section}{Section}{Sections}
\Crefname{table}{Table}{Tables}
\crefname{table}{Tab.}{Tabs.}

\def\cvprPaperID{*****} % *** Enter the CVPR Paper ID here
\def\confName{CVPR}
\def\confYear{2023}

\newcommand\blfootnote[1]{%
  \begingroup
  \renewcommand\thefootnote{}\footnote{#1}%
  \addtocounter{footnote}{-1}%
  \endgroup
}

\begin{document}

% \title{Black Box Perturbation Attack For Pre-trained Diffusion Models}

\title{
  A Pilot Study of Query-Free Adversarial Attack against Stable Diffusion
  %via Perturbations to CLIP Text Encoder
  %Query-free Perturbation Attacks Toward Diffusion Models Through Deviating CLIP Text Encoder\\
  % Achilles's Heel, Diffusion Model's CLIP: \\ A Purely CLIP Encoder-based Attack Towards Diffusion Models
  % A Vulnerable CLIP Fails a Diffusion Model: \\ A Purely CLIP Encoder-based Attack Towards Diffusion Models
  % A Diffusion Model Rots from the Head Down: \\ A Purely CLIP Encoder-based Attack Towards Diffusion Models
}

\author{Haomin Zhuang\\
South China University of Technology\\
{\tt\small semzm@mail.scut.edu.cn}
\and
Yihua Zhang\\
Michigan State University\\
{\tt\small zhan1908@msu.edu}
\and
Sijia Liu\\
Michigan State University\\
{\tt\small liusiji5@msu.edu}
}

\maketitle
%%%%%%%%% ABSTRACT

\blfootnote{\textit{The 3rd Workshop of Adversarial Machine Learning on Computer Vision: Art of Robustness at The IEEE/CVF Conference on Computer Vision and Pattern Recognition 2023}, Vancouver, Canada. Copyright 2023 by the author(s).}

\input{sec/abstract.tex}

%%%%%%%%% BODY TEXT
\input{sec/intro.tex}

\input{sec/background.tex}

\input{sec/method.tex}

\input{sec/experiment.tex}

\input{sec/limitation.tex}
\input{sec/conclusion.tex}

%%%%%%%%% REFERENCES
{\small
  \bibliographystyle{unsrt}
  \bibliography{ref_adv_OPTML,egbib}

}

\end{document}

%% file: sec/abstract.tex
\begin{abstract}
    %The field of diffusion model (DM) has seen a dramatic rise in interest in recent years, which has emerged as a popular and successful family of deep generative models.
    Despite the record-breaking performance in Text-to-Image (T2I) generation by Stable Diffusion, less research attention is paid to  its adversarial robustness. In this work, we study the problem of adversarial attack generation for Stable Diffusion
    and ask if an adversarial text prompt can be obtained even in the absence of  end-to-end model queries. We call the resulting problem `query-free attack generation'. To resolve this problem, we show that the vulnerability of T2I models is rooted in the lack of robustness of text encoders, \textit{e.g.}, the CLIP text encoder used for attacking Stable Diffusion. Based on such insight,
    we propose both untargeted and targeted query-free
    attacks, where the former is built on the most influential dimensions in the text embedding space, which we call steerable key dimensions.
    By leveraging the proposed attacks, we empirically show that only a five-character perturbation to the text prompt is able to cause the significant content shift of  synthesized images using Stable Diffusion. Moreover, we show that the proposed target attack can precisely steer the diffusion model to scrub the  targeted image content without causing much change in untargeted image content. Code is available at \url{https://github.com/OPTML-Group/QF-Attack}.

\end{abstract}

%% file: sec/intro.tex
\section{Introduction}
\label{sec:intro}

Diffusion models (DMs), the recently predominant generative modeling technique, have been  used in a wide range of
computer vision (CV) applications. Examples include Text-To-Image (T2I) generation~\cite{rombach2022high,nichol2021glide,ramesh2022hierarchical,ho2022classifier,saharia2022photorealistic},
adversarial robustness \cite{carlini2022certified,wang2022guided,nie2022diffusion}, and image reconstruction~\cite{song2021solving,abu2022adir}.
In this paper, we focus on  DM for T2I generation. The key idea following \cite{rombach2022high} is to start from a noisy input and  then iteratively refine it through a pre-trained representation network, \textit{e.g.,} CLIP (Contrastive Language-Image Pretraining)~\cite{radford2021learning} that connects texts and images. The above pipeline allows the use of various  `text prompts' (\textit{i.e.}, natural language inputs served as instructions of DM) to
effectively control the content of the synthesized images~\cite{wang2020high,ramesh2022hierarchical}.

\begin{figure}[t]
  \centering
  \begin{tabular}{ccc}
    \hspace*{-4mm}
    \includegraphics[width=.3\linewidth,height=!]{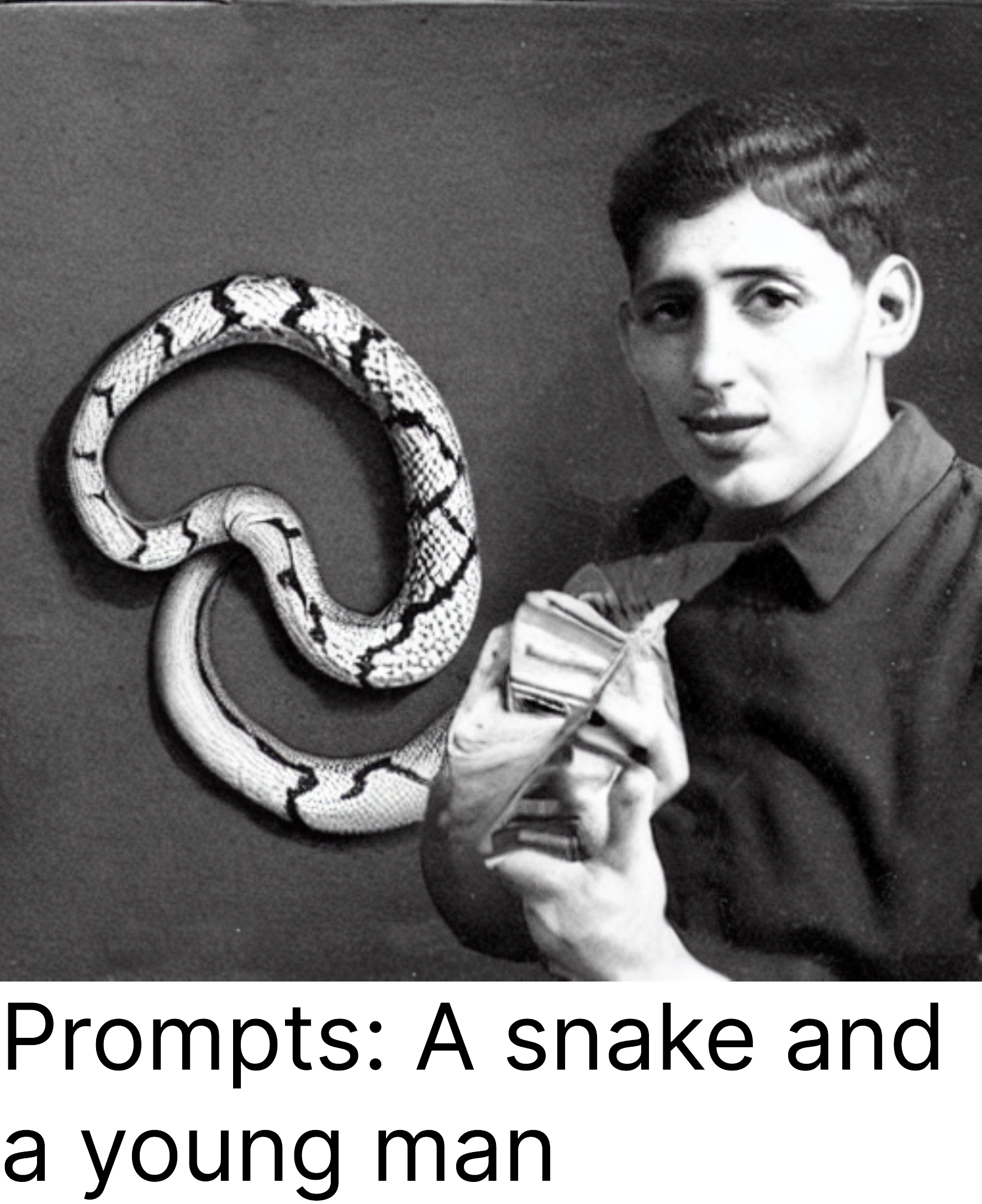}
                                                & \hspace*{-4mm}
    \includegraphics[width=.3\linewidth,height=!]{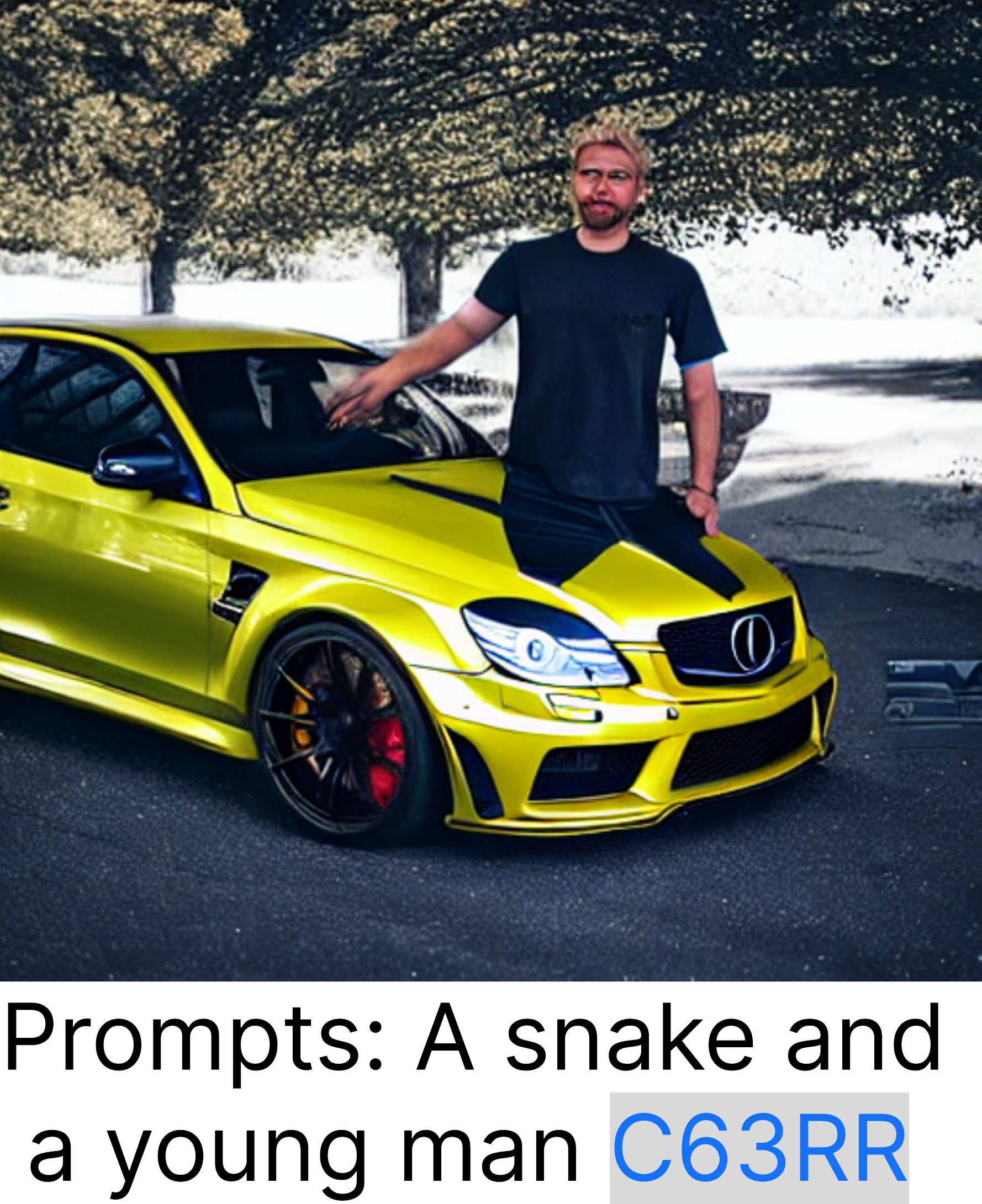}
                                                & \hspace*{-4mm}
    \includegraphics[width=.3\linewidth,height=!]{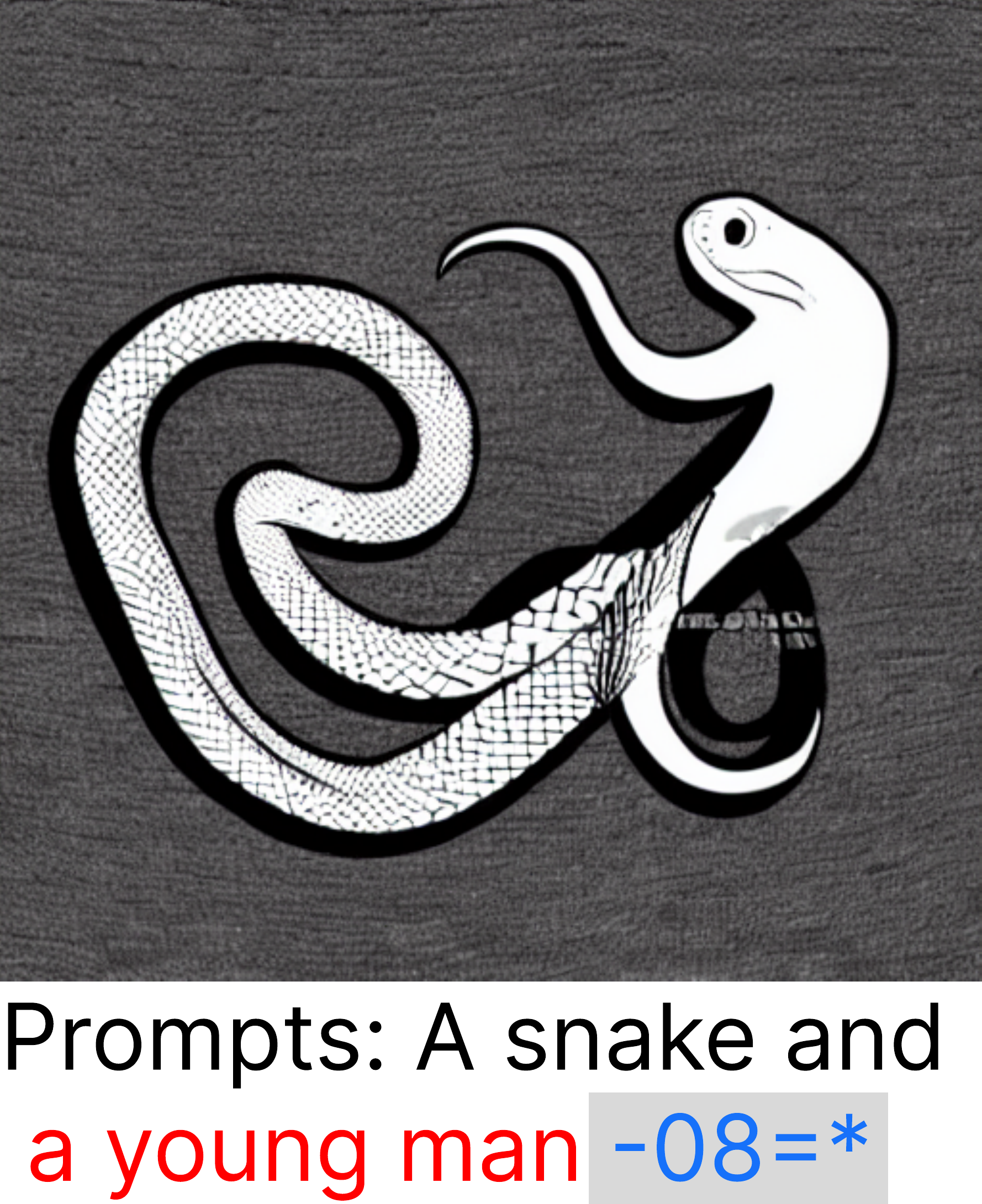}
    \vspace*{-2mm}                                                                                                                                        \\
    \hspace*{-4mm}\footnotesize{(a) No attack.} & \hspace*{-4mm}\footnotesize{(b) Untargeted attack.} & \hspace*{-4mm}\footnotesize{(c) Targeted attack.}
  \end{tabular}
  \vspace*{-2mm}
  \caption{\footnotesize{An illustration of our attack  method against Stable Diffusion. The generated perturbations are highlighted in \textcolor[RGB]{22,113,250}{blue}. The targeted attack aims to erase the image content related to   `young man' highlighted in \textcolor{red}{red}. All the images are generated from the same seed.}}
  \label{fig: teaser}
\end{figure}

\input{FigureTex/text_cos}

However,  several   works \cite{mao2022understanding,Fort2021CLIPadversarialstickers,galindounderstanding,noever2021reading} showed that  adversarial perturbations (in terms of small textual/visual input perturbations \cite{goodfellow2014explaining,wang2021adversarial}) can significantly impair the performance of a CLIP model, and thus induce the adversarial robustness concern of its downstream applications. Inspired by the above, our interest in this paper is to investigate the adversarial robustness of T2I generation using CLIP-based DMs, \textit{i.e.}, Stable Diffusion~\cite{rombach2022high} in this work. In particular, we ask:
% recent work~\cite{mao2022understanding}
% % \hl{(HM: This work focuses on perturbation attacks on images and the develop robust visual prompt. There is no other paper proposing the robustness issue on CLIP text encoder.)}
% shows that adversarial perturbations can significantly impair the performance of the CLIP models, which poses great threats to the quality and liability of its downstream tasks. Therefore, it remains elusive if the pretrained CLIP text encoder-enabled DMs, such as Stable Diffusion Model~\cite{rombach2022high} and DALL-E 2~\cite{ramesh2022hierarchical}, are robust against adversarial attacks or not. To this end, we ask:
\begin{tcolorbox}[before skip=0.2cm, after skip=0.2cm, boxsep=0.0cm, middle=0.1cm, top=0.1cm, bottom=0.1cm]
  \textit{\textbf{(Q)} Can we generate adversarial perturbations against T2I models in a query-free regime?}
\end{tcolorbox}

Adversarial attacks (also known as adversarial perturbations or   examples) that can cause models' erroneous prediction  have introduced   immense research efforts in both vision and language domains \cite{szegedy2013intriguing,goodfellow2014explaining,jiang2020robust,wang2021adversarial,hou2022textgrad}. A few recent attentions were also paid on T2I models  \cite{maus2023adversarial,wen2023hard} as different from ordinary vision or language models, the latter adopts  a natural language prompt to influence
its imagery output. The controllability and flexibility  of adjusting text prompts provide a new way to interact with a model. In \cite{maus2023adversarial},  model query-based adversarial attacks were proposed for T2I models. Yet, this work calls for many model queries ($10000$ queries per attack) to find a successful adversarial prompt, \textit{e.g.}, using 4 newly generated words integrated into  the original textual input. By contrast, our work focuses on \textit{query-free} attack generation and the perturbation is constrained to only five characters. In \cite{wen2023hard}, a gradient-based optimization was proposed to generate proper prompts that can match given images or sentences. Although it also demonstrates the controllability of image outputs by textual input prompts, little attention was paid to  adversarial robustness.
%On the contrary, our goal is to generate an extra word perturbing given sentences.

% To the best of our knowledge, the main research focus in this work \textbf{\textit{(Q)}} still remains an open question in the existing literature. The most relevant work to ours is \cite{maus2023adversarial}, which highly relies on the the heavy queries on the DM ($10000$ queries per attack) and allows attack to append up to $4$ words as perturbations. In contrast, the attack method proposed in this work does not necessitate any queries on the DM directly. Instead, our work shows that an effective attack on DMs can be generated solely based on perturbing the CLIP embeddings with a much smaller perturbation budget (as short as a 5-character word).

% aims to shift the output images to expected classes by introducing extra 4 words in the original sentences searched by 10,000 queries. 
% In contrast, our perturbation goal is to shift output images to irrelevant content with the original sentence by adding a 5-character word to perturb CLIP embeddings, where query is not needed. \YH{Emphasize the difference between this work and ours. Make it crystal clear and specific. (Done)}

% {To address \textbf{\textit{(Q)}}, we need to (1) design an effective attack paradigm to significantly alter the output of the pretrained CLIP encoders and (2) make a in-depth study on the specific text embedding learned by the CLIP encoder and the behavior of the corresponding DMs. 
% Specifically, our \textbf{contributions} are unfolded below.}

To be specific, our \textbf{contributions} are unfolded below.

%\SL{[I do not think you are showing the vulnerability of DMs. You actually show the vulnerability of text-to-image generation models?]}

\ding{172} We develop a query-free adversarial attack generator for T2I DMs. We show that a five-character perturbation, determined by   text embeddings of CLIP,  is able to significantly alter the content of DM-synthesized image     (see \textbf{Fig.~\ref{fig: teaser}b} for an illustration).

% \ding{172} We reveal the vulnerability of the DMs (diffusion models) to adversarial attacks and develop an effective attack generation method, that can effectively mislead DMs with a five-character perturbation by solely  its CLIP embeddings (see Fig.\ref{fig:apple} for illustrations). 

\ding{173} We provide an analysis of the correspondence between the semantics of synthesized images and  the  embeddings of CLIP. The obtained insight further drives us to  develop   a controllable ``targeted'' attack,
where the perturbations can be refined to steer the DM's output (see \textbf{Fig.~\ref{fig: teaser}c} for an illustration).

\ding{174} We empirically show the effectiveness of our proposal across three attack implementation methods on a variety of text-image pairs. In particular, we demonstrate that the both the proposed untargeted and targeted Query-Free Attack can successfully alter the output image content using only a 5-character prompt perturbation. This achievement is also reflected by the significantly reduced CLIP scores of the outputs.

%% file: FigureTex/text_cos.tex
\begin{figure*}[htb]
    \centering
    \includegraphics[scale=0.6]{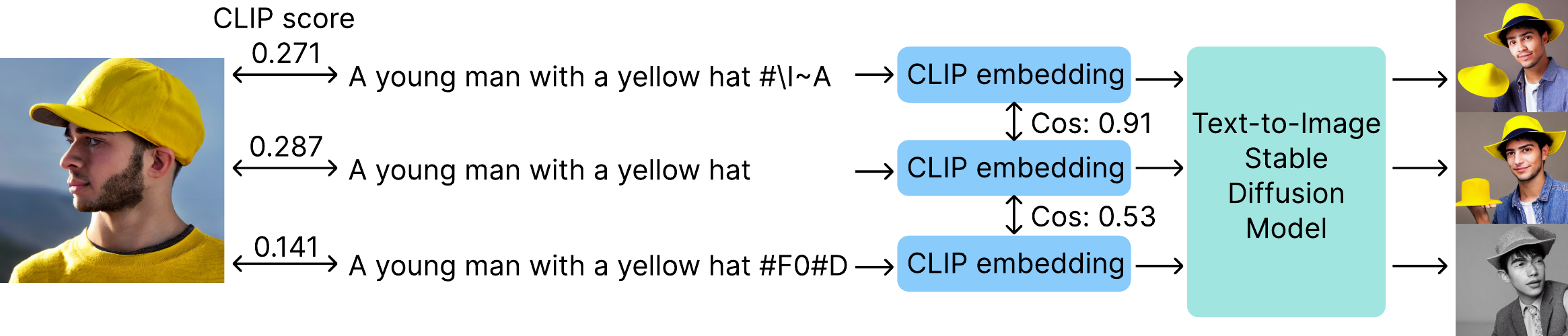}
    \caption{Illustration of   robustness issue in CLIP text encoder for image generation. CLIP score~\cite{gal2022image,wen2023hard} measures the similarity between the image-text pair provided by the CLIP model, while `Cos' measures the cosine similarity between two CLIP embeddings.}
    \label{fig:text_cos}
\end{figure*}

%% file: sec/background.tex
\section{Related Work}
% This section we introduce the overview of CLIP and the robustness issue of its text encoder. Then we introduce recent work about image output manipulation.

\paragraph{{Adversarial attacks.}}
Adversarial attacks  typically deceive DNNs by integrating  carefully-crafted tiny perturbations  into input data \cite{goodfellow2014explaining,carlini2017towards,madry2017towards,croce2020reliable,xu2019structured,chen2017ead,xiao2018spatially, liu2018signsgd, chen2017zoo, andriushchenko2020square, brendel2017decision, cheng2019sign, chen2020rays}.
Based on how an adversary interacts with the  victim model, adversarial attacks can be categorized into white-box attacks  (with full access to the victim model based on which attacks are generated) and black-box attacks  (with access only to the victim model's input   and output).
The former typically leverages the local gradient information of the victim model to generate attacks, \textit{e.g.}, \cite{goodfellow2014explaining, carlini2017towards, madry2017towards,salman2023raising}, while the latter takes input-output model queries  for attack generation; Examples include  score-based attacks (\textit{e.g.}, \cite{liu2018signsgd, chen2017zoo, andriushchenko2020square}) and decision-based attacks (\textit{e.g.}, \cite{brendel2017decision, cheng2019sign, chen2020rays}).
In this work, we assume that the adversary has access to the CLIP text encoder but can be blind to the diffusion model for image generation.
% \HM{Shall we explain the setting here like: Because recently widespread DMs~\cite{rombach2022high,ramesh2022hierarchical,saharia2022photorealistic} utilize pre-trained text encoders (\textit{e.g.}, CLIP and T5), which is open-source and easily accessed by the public.}
Our goal is to design an adversarial attack to fool the stable diffusion model without executing the diffusion process, which would take  a high model query and computation cost.
Thus, we term our proposal the `query-free attack'.
%Given the publicity of CLIP, our proposal can also be regarded as a query-free black-box attack on DMs. 
% \SL{[needs discussion.]}

\paragraph{{Prompt perturbations in vision-language models.}}
Recent studies~\cite{daras2022discovering,maus2023adversarial,wen2023hard} have explored the over-sensitivity of  text-to-image diffusion models   to prompt perturbations in the text domain. The adversarial robustness problem of CLIP was also studied in
\cite{mao2022understanding,Fort2021CLIPadversarialstickers,galindounderstanding,noever2021reading}, such as the design of
% perturbation on prompts to manipulate TTI DMs in text space. Gibberish words~\cite{daras2022discovering} are found to possess specific meanings, such as ``Apoploe vesrreaitais'' presenting birds in DALLE-2. In \cite{maus2023adversarial}, a black-box attack is proposed to shift given sentences to output particular objects with four extra words in the token table. 
%
% \paragraph{Robustness of CLIP.}
% A spectrum of studies exist in the literature, that demonstrate that CLIP's performance can be significantly impacted 
%by 
imperceptible pixel perturbations~\cite{mao2022understanding,Fort2021CLIPadversarialstickers} and attacks in the image frequency domain  \cite{galindounderstanding}.
%, and spanning attacks~\cite{noever2021reading,Fort2021CLIPadversarialstickers}.
%In the meantime, recent literature~\cite{zhou2022learning,shu2022test} shows that the quality of the text prompting can greatly influence the performance of the downstream tasks. Yet,
Yet, the previous studies focused on perturbations to image inputs of CLIP,
it lacks investigation into how the textual perturbation to CLIP can influence the T2I diffusion model.

%% file: sec/method.tex
\section{Our Proposal}

\paragraph{Problem statement.}  
In this section, we first present an overview of Stable Diffusion, and then introduce our objective to generate small perturbations on the textual inputs so as to maneuver the DM's synthesized images.

We choose   Stable Diffusion   as the victim T2I DM model due to its popularity and availability as an open-source model. 
In Stable Diffusion, the DM  denoises images in latent space and utilizes a cross-attention mechanism to guide the denoising process.
In addition,  text inputs (or textual prompts) are processed by the CLIP's text encoder to generate text embeddings and are then sent to the cross-attention layer in the denoising network. This eventually determines the synthesized images based on the CLIP's textual embeddings and the selected random seed of the initial noisy pixels. However, as exemplified in \textbf{Fig.\,\ref{fig:text_cos}}, small perturbations on the text input of CLIP can lead to different CLIP scores \cite{gal2022image,wen2023hard}, given by the values of cosine similarity of every text-image input pair. This is because of the sensitivity of the CLIP's text embedding to text perturbations. Based on that, \underline{we ask}: Can we generate an adversarial textual prompt by leveraging the lack of robustness of the CLIP's text encoder so as to fool the DM-based image generator in Stable Diffusion?

% deviate the CLIP embedding and affect the output images in the T2I Stable Diffusion Model, leading to robustness issues. Therefore, we rise the problem of how to generate small perturbations (e.g., adding a five-character word) on input sentences without querying diffusion models to impact the output images in the T2I DM.

% We begin with introducing the background of Text-To-Image diffusion models. Based on that, we reveal the possible robustness issues on the design and define the problem of interest in how to generate perturbation words. Specifically, we choose a popular diffusion model-Stable Diffusion Model~\cite{rombach2022high} as the victim model. The Stable Diffusion Model introduces denoising in latent space and guides the denoising process by cross-attention mechanism. The input text $\boldsymbol{y}$ is processed to CLIP embedding $\tau_\theta(\boldsymbol{y})$ by the CLIP text encoder $\tau_\theta$ and sent to the cross-attention layer. In this Text-To-Image generation process, the output images are decided by the selected random seed of initial noise and the CLIP embedding of the input sentence. As the robustness issue is shown in Fig.~\ref{fig:text_cos}, the small perturbation on text can deviate CLIP embedding and affect the output images. 

% We rise such a problem that how to generate small perturbations (e.g., a five-character word) on sentences to impact the output images in the Text-To-Image generation process without querying diffusion models.

% \SL{Introduce the victim model a bit used in this paper.}

% \SL{[How to define adversarial perturbations?]}

\paragraph{Attack model.}
We assume that the adversary has access to the trained text encoder   of the CLIP model, and can perturb the textual prompt of the trained State Diffusion model using an additional word within a   \textit{five-character} length. Let $\tau_\theta(\mathbf x)$  denote the text encoder of CLIP with parameters $\theta$ evaluated at the textual input $\mathbf x$. And we denote by $\mathbf x^\prime$ the perturbed textual prompt used as the input of   Stable Diffusion.  
We then define the \textit{attacker's objective} by minimizing the cosine similarity  between the text embeddings of $\mathbf x$ and $\mathbf x^\prime$. This leads to the following attack generation problem 
\begin{equation}
\label{eq:3}
\displaystyle \min_{\boldsymbol{x}^\prime } ~ \text{cos} ( \tau _\theta (\boldsymbol{x}) , \tau _\theta (\boldsymbol{x}^\prime ) ),
\end{equation}
where $\mathrm{cos}$ refers to the cosine similarity metric. 

Despite the simplicity of the attack generation in \eqref{eq:3}, we will show that it can be used to attack State Diffusion in a targeted way effectively. More importantly, the generation of the perturbed input $\mathbf x^\prime$ no longer relies on the optimization over the diffusion model, and can thus be computationally efficient. This is in contrast to  \cite{wang2021adversarial}, which requires $10000$ queries to diffusion model for generating a single attack. 
Since no attack intention is  specified in \eqref{eq:3}, we call the resulting attack an `\textit{untargeted attack}'.

\paragraph{Attack methods.}
Since problem \eqref{eq:3} is differentiable, various optimization methods can be adopted for attack generation.  Inspired by the previous studies on   adversarial attacks in the language domain,  we consider the following attack methods.

%We propose three well-study methods to generate perturbation words and evaluate their effectiveness on the T2I Stable Diffusion Model.

\textbf{PGD attack.}  Similar to the PGD (projected gradient descent) attack \cite{madry2017towards} in the image domain,   the PGD attack in the language domain  has also been developed  \cite{hou2022textgrad,srikant2021generating}. 
The key idea is to formulate the textual perturbation problem as a token selection problem (over a set of token candidates) when a token site is determined for perturbation. Our experiments follow the PGD implementation in \cite{hou2022textgrad} to solve problem  \eqref{eq:3}. 

% is applied in Nature Language Processing by \textsc{TextGRAD}~\cite{hou2022textgrad}, which is incorporated in our work. It regards the choice of candidates as a selection problem. We create a character table that consists of digits, letters, and common symbols, where characters are selected from this table. A selection vector chooses characters in each position through Gumble Softmax and is updated by the loss in Eq.~\eqref{eq:3}.

\textbf{Greedy search.} Different from the above PGD attack, we next consider a heuristics-based perturbation strategy.  We conduct a greedy search on the  character candidate set to select the top 5  characters (used for textual perturbation), which can reduce the loss of \eqref{eq:3} to the maximum extent. 

% is the most simple method to search for characters to form a word added to the end of the sentence. The algorithm operates by selecting the character that results in the minimal loss in Eq.~\eqref{eq:3}, and iteratively repeating this process until a maximal length $m$ of the generated input is reached.

\textbf{Genetic algorithm.}
We follow \cite{holland1992genetic} to generate  a population of perturbation candidates and use the loss of \eqref{eq:3} to evaluate the quality of each candidate. In each iteration, the genetic algorithm calls genetic operations such as mutation  to generate new candidates. The process continues until the number of generations is met. 
%The advantage of the genetic algorithm is that it can explore a broader search space and find globally optimal solutions without gradient.

\input{FigureTex/flooding}
\paragraph{Targeted  attack and steerable key  dimensions.}
In what follows, \underline{we investigate} if the  attack generated by \eqref{eq:3} can be further \textit{refined}  towards a \textit{targeted} attack purpose, \textit{e.g.}, the intention of removing the `young man'-related image content from the original image in Fig.\,\ref{fig: teaser}-(c) vs. (a).
To this end, we propose a new concept termed \textit{steerable key dimensions} in the text embedding space, along which the attack generator can be guided to design customized textual perturbations.  By constraining the perturbations on these steerable key dimensions, we can improve the likelihood of image generation following the  adversary's intention.

To be specific, we first generate a sequence of augmented sentences $\{ s_i \}_{i=1}^n$ that reflect the adversary's  intention, \textit{e.g.},  the sentences centered on `a young man' in Fig.\,\ref{fig: teaser}-(a). The generation of $\{ s_i \}_{i=1}^n$ can be realized using \textit{e.g.}, ChatGPT by requesting `Generate $n$ simple scenes and end with ``and a young man'' without extra words'. {Two examples are $s_1 = \text{`A bird flew high in the sky and a young man'}$ and $s_2 = \text{`The sun set over the horizon and a young man'}$ with $n = 2$.} Next, we perturb $\{ s_i \}_{i=1}^n$  by \textit{removing} the adversary's intention-related sub-sentence (\textit{i.e.}, `a young man' in Fig.\,\ref{fig: teaser}-(a)). This results in a modified sequence $\{ s_i^\prime \}_{i=1}^n$; For example,  {$s_1^\prime = \text{`A bird flew high in the sky'}$ and $s_2^\prime = \text{`The sun set over the horizon'}$}. 
We then  obtain the corresponding CLIP embeddings $\{ \tau_\theta (s_i) \}_{i=1}^n$ and $\{ \tau_\theta (s_i^\prime) \}_{i=1}^n$. 
As a result, the text embedding difference $\boldsymbol{d}_i = \tau_\theta (s_i)-\tau_\theta (s^\prime_i)$ can characterize the \textit{saliency} of the   adversary's intention-related sub-sentence in the text embedding space. 
For $n$ such   difference vectors $\{ \mathbf d_i \}_{i=1}^n$, we determine the  \textit{steerable key dimensions} for targeted attack generation by identifying the most influential dimensions in the difference vectors $\{ \mathbf d_i \}_{i=1}^n$. The dimension influence is given by a majority vote of $\{ \mathbf d_i \}_{i=1}^n$ along each dimension. That is, $I_j = 1$ (\textit{i.e.}, the indicator of the $j$th dimension being influential), if $|\textstyle \sum_{i=1}^{n}\text{sign}({d}_{i,j})|>\epsilon  n$, where $\mathrm{sign}$ is the sign operation, ${d}_{i,j}$ is the $j$th entry of $\mathbf d_i$, and $\epsilon < 1  $ is a threshold to pick the most influential dimensions. As a result, the binary vector $I$ encodes the selected key dimensions. By integrating $ I$ into \eqref{eq:3}, we obtain the key dimensions-guided targeted attack generation 
\begin{equation}
\label{eq:cos_mask}
   \displaystyle \min_{\boldsymbol{x}^\prime } ~~\text{cos} ( \tau_\theta (\boldsymbol{x})\odot I,\tau_\theta (\boldsymbol{x}^\prime )\odot I),
\end{equation}
where $\odot$ is the element-wise product. Note that problem \eqref{eq:cos_mask} can be similarly solved as  \eqref{eq:3} using the attack methods introduced before. Fig.\,\ref{fig: teaser}-(c) shows an example   of  using prompt perturbations generated by the proposed targeted attack to erase the image content related to  `a young man'.

%% file: FigureTex/flooding.tex
% \begin{figure}[t]
%     \centering
%     \subfloat[Prompt ``A car on the side of the street'']{\includegraphics[scale=0.15]{figs/car/orgin.png}}\hspace{5mm}
%     \subfloat[Prompt ``A car on the side of the flooded street'']{\includegraphics[scale=0.15]{figs/car/flooded.png}}\hfill
%     \subfloat[Prompt ``A car on the side of the street flooding'']{\includegraphics[scale=0.15]{figs/car/flooding.png}}\hspace{5mm}
%     \subfloat[Manipulate text embedding on steerable dimensions]{\includegraphics[scale=0.15]{figs/car/key.png}}
%     \caption{The image generated from the prompt "A car on the side of the street" is manipulated through two methods: adding words in the input space and adding a
%         \textit{text diff} vector in steerable dimensions $\boldsymbol{l} \odot \boldsymbol{h}$. In the latter method, the difference vector is added to the CLIP embedding of the original prompt after multiplying it with a scale.  \SL{[remove this figure.]}
%         % \YH{Move this figure to the page where it is called for the first time. (Done)}
%         }
%     \label{fig:Flooding}
% \end{figure}

%% file: sec/experiment.tex
\section{Experiment}

\input{FigureTex/res_pa}
\input{FigureTex/res_rpa}

\subsection{Experiment setups}

\paragraph{Model setup.} 
Throughout the experiments, we use Stable Diffusion model v1.4~\cite{rombach2022high} as the victim model for image generation.  %to generate $512\times 512$. 
%images with the classifier free guidance scale of 7.5 and 50 inference steps.
The proposed query-free attack has the access to   the CLIP model (ViT-L/14) that shares the same text encoder as     Stable Diffusion. The CLIP model is trained on a dataset containing text-image pairs 
%in websites and commonly-used pre-existing image datasets such as YFCC100M~
\cite{thomee2016yfcc100m}. 
%and the Stable Diffusion Model is trained on multiple datasets such as laion2B-en~\cite{schuhmann2022laion}. 
% \YH{How is the model pretrained? Any source or references for the model checkpoints?} \HM{Done!}

% We utilize Stable Diffusion Model version v1.4 to generate $512\times 512$ images with classifier free guidance scale of 7.5 and 50 inference step. The version of CLIP is ViT-L/14 which has the same text encoder as Stable Diffusion Model. 
\paragraph{Attack implementation.}
When implementing the PGD attack method, we set the base learning rate  by $0.1$ and the number of PGD steps by 100. When implementing the genetic algorithm, we set the number of generation steps, the number of candidates per step, and the mutation rate by $50$, $20$, and $0.3$. When implementing the targeted attack, we use ChatGPT~\cite{chatGPT2023} to generate $n = 10$ sentences to characterize the steerable key dimensions and set $\epsilon = 0.9$ to determine the influence mask $I$ in \eqref{eq:cos_mask}. In addition, we utilize ChatGPT generating $20$ prompts forming an input text dataset for the quantization by requesting `Generate 20 simple scenes for text-to-image generative model'.

% For PGD attack, the base learning rate is set to $0.1$. In the Genetic algorithm, the maximum generation step is set to 50, the selected candidates number in each generation is set to 20, and the mutation rate is set to 0.3. For the targeted perturbation attacks, we set $\epsilon = 0.9$ in \eqref{eq:cos_mask}. 
% % In this setting, the $\boldsymbol{h}$ \YH{in Eq.(?)} refers to the mask of dimensions in which all \textit{text diff} share the same direction.
% To mitigate bias, we use ChatGPT~\cite{chatGPT2023}, a large language model, to generate 20 simple prompts for text-to-image generation models forming a text input dataset by requesting `Generate 20 simple scenes for text-to-image generative model' 
% % \YH{why? Better to put a reference here}.

\paragraph{Evaluation metrics.}
%To demonstrate the effectiveness of our proposed query-free attack,
In addition to different  implementations of our proposed query-free attack (\textit{i.e.}, PGD, Greedy, and Genetic methods), 
we also introduce a  baseline that randomly generates random five-character prompt perturbations, termed Random. 
We evaluate the effectiveness of an attack using 
 the CLIP score~\cite{gal2022image,wen2023hard} to characterize the similarity between the text input and the generated image. A lower CLIP score represents a lower semantic correlation between the generated image and the input text, indicating the higher effectiveness of the attacking method. 
 To quantify the CLIP score between the targeted objects and images generated in the targeted attack setting, we utilized the template sentence `This is a photo of'~\cite{radford2021learning} as text input to measure text-image pair similarity.
The CLIP score reported for each method will be averaged over 20 prompts, based on each of which 10 images will be generated.  
%10 generated images given a prompt.
%During the evaluation, for all the methods and baselines, ten potential perturbation words are proposed for each prompt and the best one is selected.

\subsection{Experiment results}

\begin{table}[tb]
    \caption{CLIP scores~\cite{gal2022image,wen2023hard} comparison of images generated with different methods.  CLIP scores are used to indicate the similarity between the generated images and the embeddings of the corresponding text prompts.
    For each method, the CLIP scores reported below are averaged over $20$ prompts and $10$ images per prompt. In particular, the scores calculated based on the original sentences and output images are adopted for the untargeted attack and based on the targeted content and output images for the targeted setting.
    The lowest (best) score in each row is in \textbf{bold} and the results in the form $a$\footnotesize{$\pm b$} denote the mean value $a$ and the standard deviation $b$.
    % The scores indicate the similarity between the generated images after attacks and the original sentences. The attacks include: no attack, random selecting attack (Random), greedy attack (Greedy), genetic attack (Genetic), and PGD attack (PGD).
    }
    \label{tab: exp_main}
    \centering
    \resizebox{\columnwidth}{!}{%
        \begin{tabular}{c|c|c|c|c|c}
            \toprule[1pt]
            \textbf{Method}:
            & \textbf{No Attack} & \textbf{Random} 
            & \textbf{Greedy} & \textbf{Genetic} 
            & \textbf{PGD}   \\ \midrule
            \multicolumn{6}{c}{Untargeted Attack} \\ \midrule
            Score:       
            & 0.277\footnotesize{$\pm 0.022$}     
            & 0.271\footnotesize{$\pm 0.021$}  
            & \cellcolor{Gray} 0.255\footnotesize{$\pm 0.039$}  
            & \cellcolor{Gray}\textbf{0.203}\footnotesize{$\pm 0.042$}  
            & \cellcolor{Gray} 0.226\footnotesize{$\pm 0.041$} \\ \midrule
            \multicolumn{6}{c}{Targeted Attack} \\ \midrule
            Score:       
            & 0.229\footnotesize{$\pm 0.03$}     
            & 0.223\footnotesize{$\pm 0.037$}  
            & \cellcolor{Gray} 0.204\footnotesize{$\pm 0.037$}  
            & \cellcolor{Gray} \textbf{0.186}\footnotesize{$\pm 0.04$}   
            & \cellcolor{Gray} 0.189\footnotesize{$\pm 0.041$} \\
            \bottomrule[1pt]
        \end{tabular}
    }
\end{table}

% \begin{table}[tb]
%     \caption{CLIP scores~\cite{gal2022image,wen2023hard} for generated images under different attacks. The scores indicate the similarity between the generated images after attacks and the target objects. The attacks include: no attack, random selecting attack (Random), greedy attack (Greedy), genetic attack (Genetic), and PGD attack (PGD).}
%     \label{table:result2}
%     \centering
%     \scalebox{0.9}{
%         \begin{tabular}{l|c|c|c|c|c}
%             \toprule[1pt]
%             \multicolumn{1}{c|}{Attack} & No Attack & Random & Greedy & Genetic & PGD   \\ \midrule
%             Score                        & 0.2123     & 0.2071  & 0.215  & 0.1768   & 0.1723 \\ \bottomrule[1pt]
%         \end{tabular}
%     }
% \end{table}

\paragraph{Query-free attack can successfully alter the image output of Stable Diffusion using only a 5-character prompt perturbation.}
%先从图片说起，因为图片和输入发生了偏离。偏离包括遗忘原句中的一些词和新添加了一些词。举例，单车被perturbation word扰动后，从图片消失。其他例子也相似，这说明了攻击的成功。除此之外，在量化的结果中，和baseline相比有着显著的下降，也证明了我们攻击的有效性。在横向的对比中，greedy方法的作用有限，genetic有最好的效果。说明genetic可以在不同generation的迭代中找到可以持续降低相似度的字符组合。
 In \textbf{Fig.\,\ref{fig:res_pa}}, we present examples of text-to-image generation \textit{with} and \textit{without} suffering prompt perturbations generated by the untargeted, genetic algorithm-based query-free attack. The 5-character prompt perturbation is highlighted in \textcolor[RGB]{22,113,250}{blue} with \textcolor{gray}{gray}  background. 
As we can see, the proposed attack can significantly alter the content of the original image  produced by Stable Diffusion. For example, in Fig.\,\ref{fig:res_pa}-(a), the perturbation `E\$9\textbackslash' ' drives the model to generate images far from the true topic `bicycle'.
The same observation can also be drawn from examples in  Fig.\,\ref{fig:res_pa}-(b)-(h).
This implies that the attack against the text embedding remains effective in manipulating the output of text-to-image generation.

Similar to Fig.\,\ref{fig:res_pa}, 
  \textbf{Fig.\,\ref{fig:rpa}} presents examples of targeted, genetic-based query-free attacks against Stable Diffusion. For example, in Fig.\,\ref{fig:rpa}-(a), 
  the adversary targets   perturbing `bicycle' without altering the background `brick wall' much. 
 This contrasts to Fig.\,\ref{fig:res_pa}-(a), where the image object and scene may change.   
 Another example is  Fig.\,\ref{fig:res_pa}-(b), where 
 the object `plate' is erased using the perturbed prompt but the object `apple' is retained.  We can also draw similar observations from  other examples. 
 Briefly, the targeted attack can precisely manipulate the diffusion model to avoid the targeted semantics (\textit{i.e.}, the   \textcolor{red}{red} text highlight above each image example in {Fig.\,\ref{fig:rpa}), while  can retain the irrelevant semantics (\textit{e.g.}, `{brick wall}' in Fig.\,\ref{fig:res_pa}-(a)).
 
 %while erasing the targeted ones (\textit{e.g.}, `\textcolor{red}{bicycle}', `\textcolor{red}{plate}', `\textcolor{red}{lake}', `\textcolor{red}{leaf}'). 

% To illustrate the effect of the proposed query-free attack, we first present the examples generated by the T2I Stable Diffusion model \textit{with} and \textit{without} perturbations in Fig.\,\ref{fig:res_pa} (through untargeted query-free attack) and Fig.\,\ref{fig:rpa} (through targeted query-free attack). As we can see, the model can work very well and generate images in the topics dictated by the prompt when no perturbation is planted, (see the images on the first row). However, the generated images consistently go through a dramatic content topic shift in the presence of the perturbations (highlighted in \textcolor[RGB]{22,113,250}{blue}). 
% For example, in Fig.~\ref{fig:res_pa}-(a), the perturbation `E\$9\textbackslash' ' drives the model to generate images far from the true topic `bicycle'. Besides, in Fig.\,\ref{fig:rpa},  the targeted attack is able to not only alter the topics, but also precisely manipulate the victim model to avoid the targeted semantics (highlighted in \textcolor{red}{red} above the images). In the meantime, the targeted attack achieves to retain the irrelevant semantics (\textit{e.g.}, `{brick wall}', `{apple}', `{green}{swan}', and `{green}{butterfly}'), while erasing the targeted ones (\textit{e.g.}, `\textcolor{red}{bicycle}', `\textcolor{red}{plate}', `\textcolor{red}{lake}', `\textcolor{red}{leaf}'). 

\paragraph{Query-free attack can effectively reduce the CLIP score.}
To quantify the influence of the attack in each generated text-image pair, \textbf{Table\,\ref{tab: exp_main}} presents the CLIP scores \cite{gal2022image,wen2023hard} of the image pairs generated by Stable Diffusion with and without the  attack's perturbations. 
% \HM{Here $20$ prompts in the input text dataset are evaluated.} 
In Table\,\ref{tab: exp_main}, we can make the following observations.
\underline{\textit{First}}, in the untargeted attack setting, it is clear that  different attack methods can all successfully reduce the CLIP score versus the baseline result using `No Attack' or `Random' attack strategy. Such a reduction implies a relatively low similarity between the perturbed text input and the generated image and  justifies the image content modification   observed in Fig.\,\ref{fig:res_pa}. 
Moreover, the genetic algorithm outperforms the other attack methods. This also supports the choice of genetic algorithm-based untargeted attack in Fig.\,\ref{fig:res_pa}.
%In contrast, the random perturbation hardly reduces the CLIP score in all the settings, which necessitates to carefully design the perturbation patterns. 
\underline{\textit{Second}}, in the targeted attack scenario, the PGD attack method and the genetic algorithm outperform other attack methods. 
%in finding perturbations altering the specific embeddings of the CLIP text encoder. 
We also notice that the targeted attack setting introduces additional difficulty for effective perturbation generation, evidenced by the smaller drop in the CLIP score compared to the untargeted setting.

\begin{figure}[tb]
    \centering
    \includegraphics[scale=0.1]{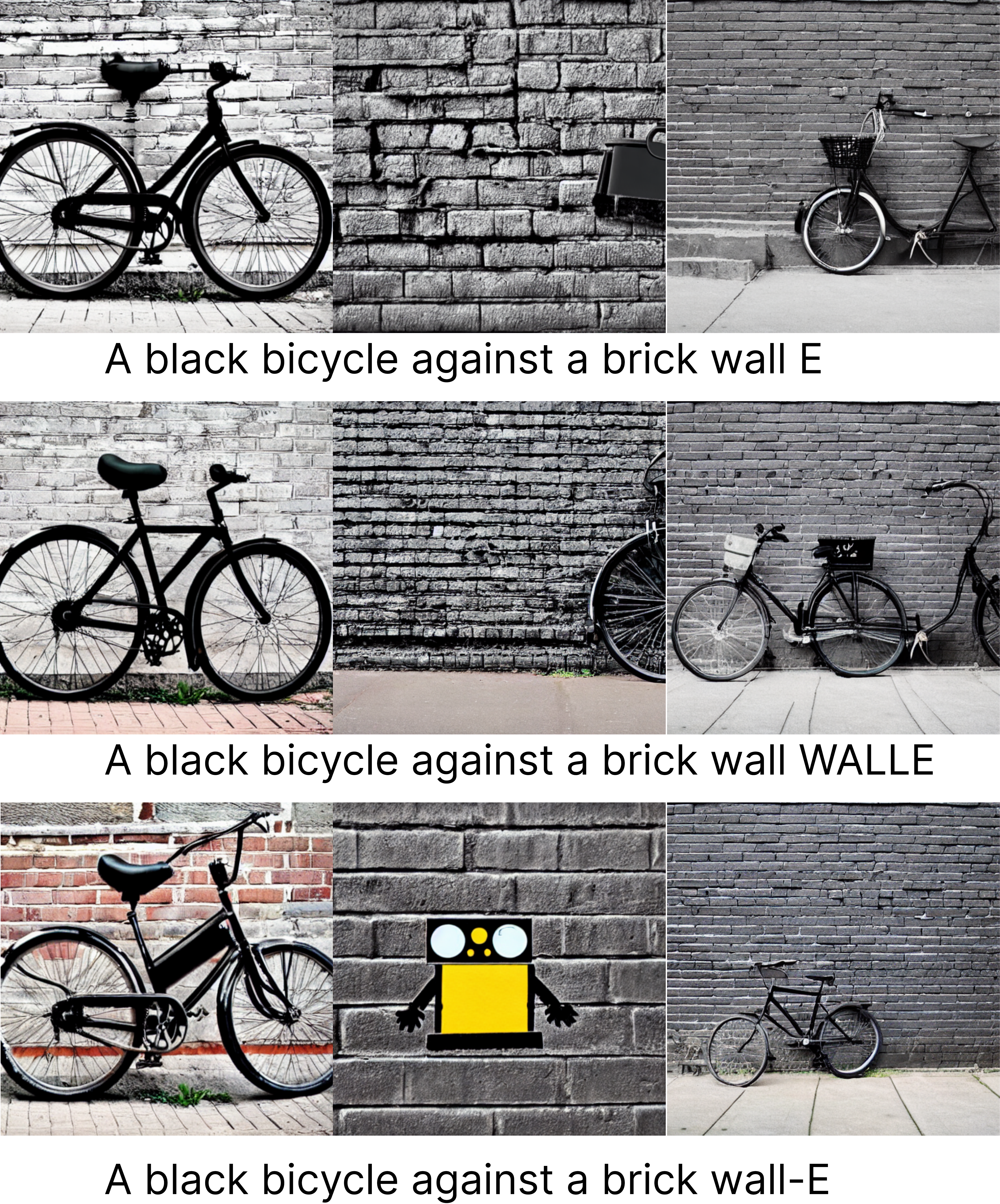}
    \caption{Ablation study for the perturbation word generating  robots in Fig.\,\ref{fig:res_pa}.}
    % \HM{original: 0.293, ours: 0.217, E:0.297, WALLE: 0.291, -E: 0.285}
    \label{fig:walle}
\end{figure}

\begin{table}[tb]
    \caption{CLIP scores~\cite{gal2022image,wen2023hard} comparison of different perturbation prompts in case study.  
    For each prompt, the CLIP scores reported below are averaged over $10$ images from the same seeds.
    The lowest (best) score in each row is in \textbf{bold}.
    }
    \label{tab: walle}
    \centering
    \resizebox{\columnwidth}{!}{%
        \begin{tabular}{c|c|c|c|c|c}
            \toprule[1pt]
            \textbf{Perturbation prompt}:
            & \textbf{None} & \textbf{`E\$9\textbackslash' '} 
            & \textbf{`E'} & \textbf{`WALLE'} 
            & \textbf{`-E'}   \\ \midrule
            
            Score:       
            &  0.293
            & \cellcolor{Gray}\textbf{0.217}  
            & 0.297
            & 0.291
            &  0.285 \\ 
            \bottomrule[1pt]
        \end{tabular}
    }
\end{table}
\paragraph{Why does the perturbation work? A case study on `WALL-E'.}
To demonstrate the effectiveness of our attacks, we conduct an ablation experiment to compare the perturbations generated by our method and a direct change in textual semantics.
%Despite the limited  vocabulary in the character table, some perturbations seem   explainable and relevant to the desired image output. 
Recall from {Fig.\,\ref{fig:res_pa}-(a)} that the generated perturbation `E\$9\textbackslash' ' appended to the sentence `A black bicycle against a brick wall' seems   related to a \textit{robot movie `WALL-E'}\footnote{\url{https://en.wikipedia.org/wiki/WALL-E}.}. We   wonder if this is due to the effect of the combination between `wall' in the original text input and the   added letter `E' in the generated perturbation   `E\$9\textbackslash' '. 
To this end, we conduct additional experiments to explicitly append the letter `E' to the end of the original text input.  \textbf{Fig.~\ref{fig:walle}} shows that simply adding `E' or  `WALLE' fails to alter the image content (see the first two rows of Fig.~\ref{fig:walle}). 
Although replacing `wall' with `wall-E' in the original sentence may produce a robot-related image, the success of such image generation remains low. This trend is also supported by the corresponding CLIP scores reported in Table~\ref{tab: walle}, where almost no change can be observed (see $0.293$ \textit{vs.} $\{0.297, 0.291, 0.285\}$).
% \HM{ as shown in Table~\ref{tab: walle} that both cause almost no change in CLIP score from 0.293 to 0.297 and 0.291 respectively}.
By contrast, the use of prompt perturbation `E\$9\textbackslash' ' in {Fig.\,\ref{fig:res_pa}-(a)} is much more effective in altering the image content (see a CLIP score drop from $0.293$ to $0.217$ in Table~\ref{tab: walle}.

%% file: FigureTex/res_pa.tex
% \begin{figure*}[t]
%     \centering
%     \includegraphics[scale=0.078]{figs/pa/res_pa.jpg}
%     \caption{Samples of perturbation attacks applied to sentences in the test dataset. The original images and perturbed images are generated from the same seeds.}
%     \label{fig:res_pa}
% \end{figure*}

\begin{figure*}[t]
    \centering
    \subfloat[\label{fig:bike}]{\includegraphics[scale=0.078]{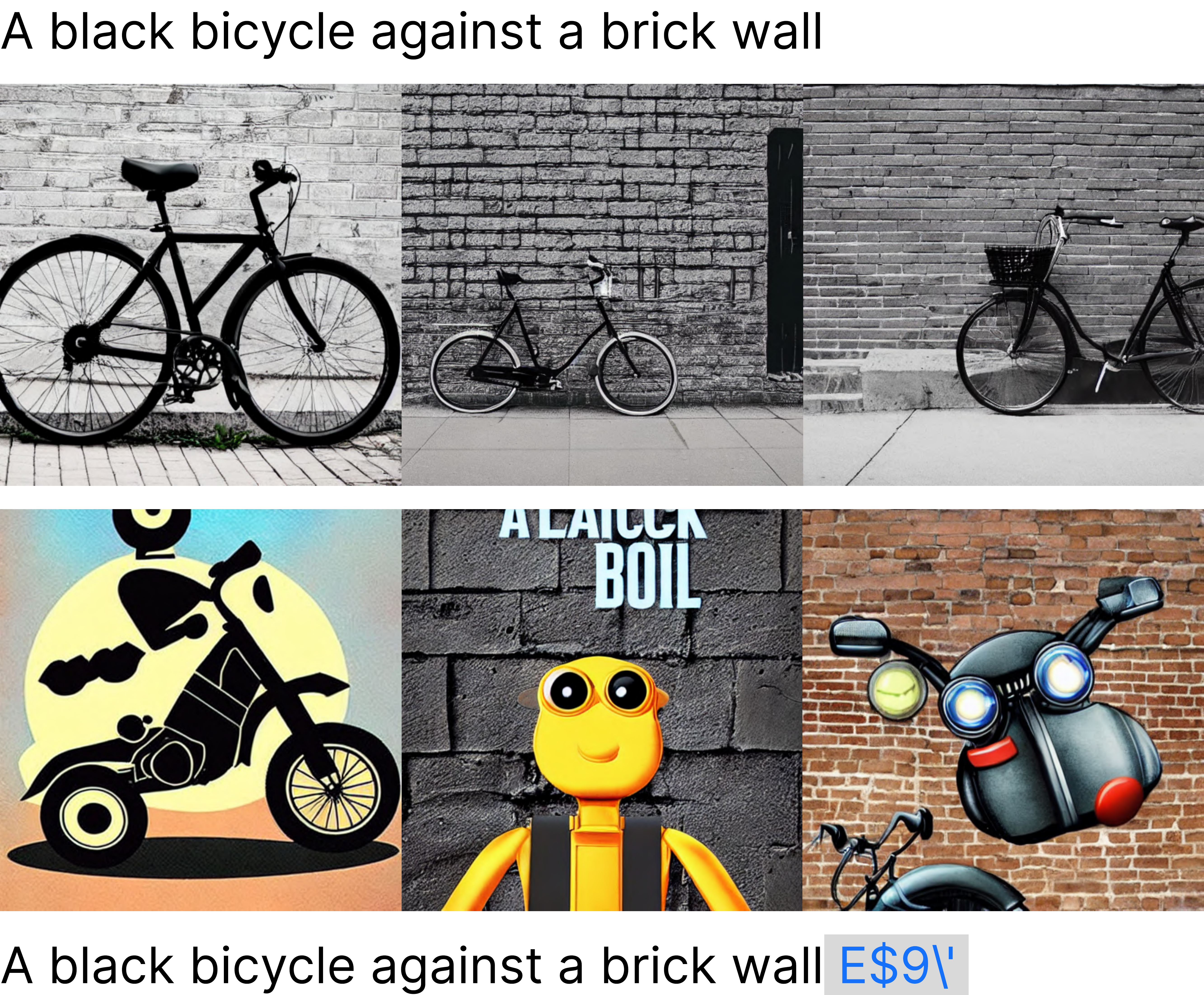}}
    \hspace{1mm}
    \subfloat[]{\includegraphics[scale=0.078]{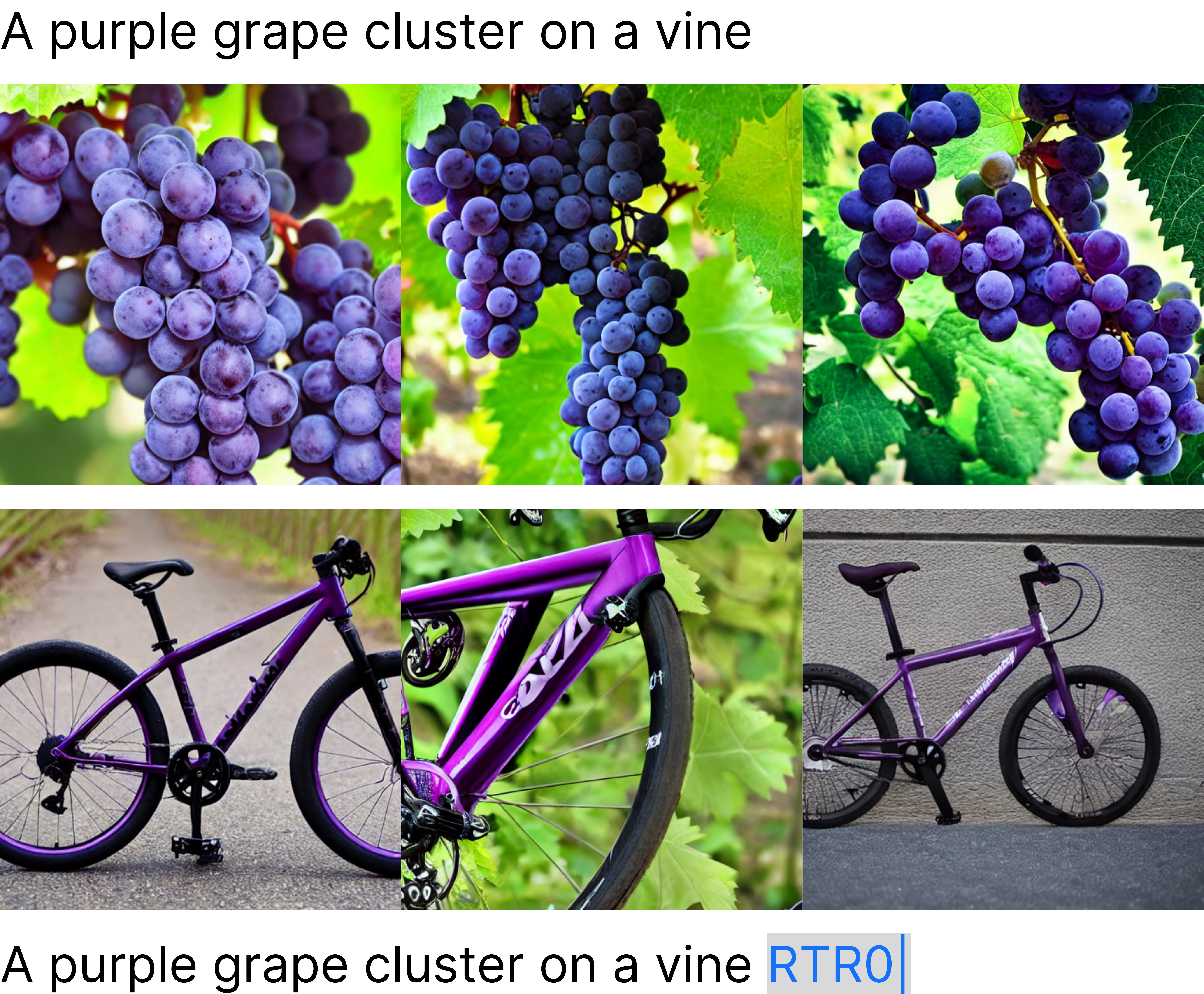}}
    \hspace{1mm}
    \subfloat[]{\includegraphics[scale=0.078]{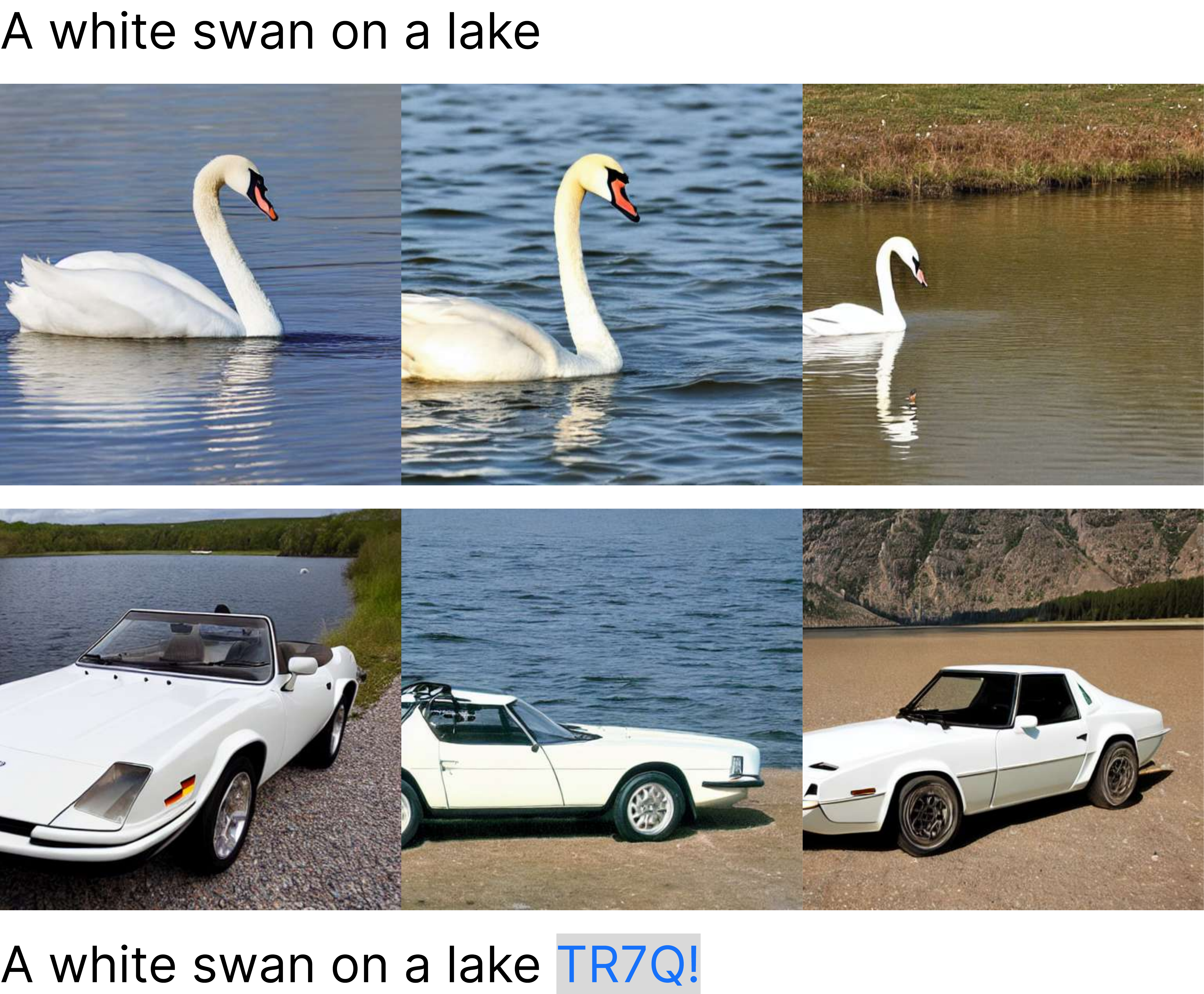}}
    \hspace{1mm}
    \subfloat[]{\includegraphics[scale=0.078]{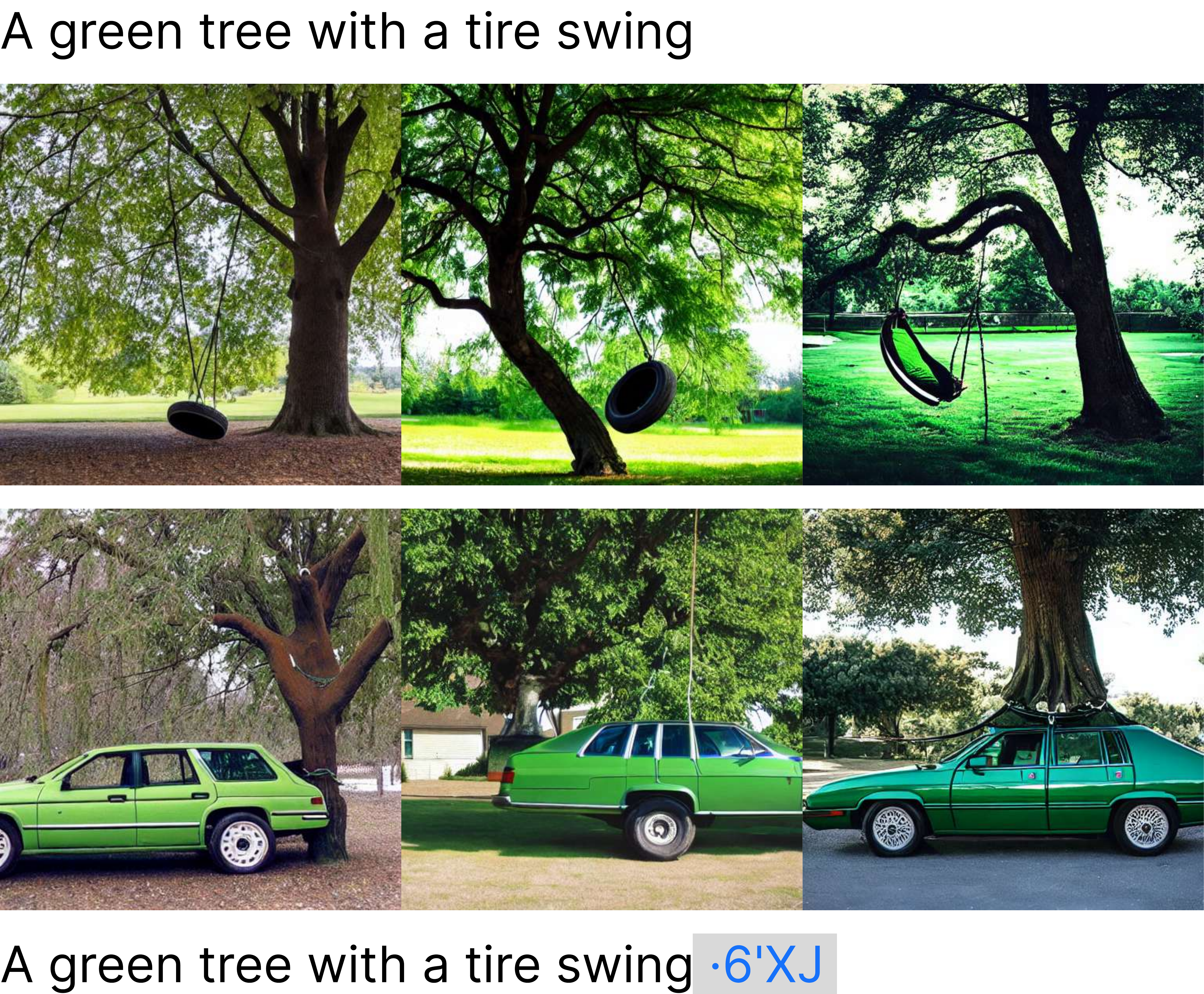}}
    \hfill
    \subfloat[]{\includegraphics[scale=0.078]{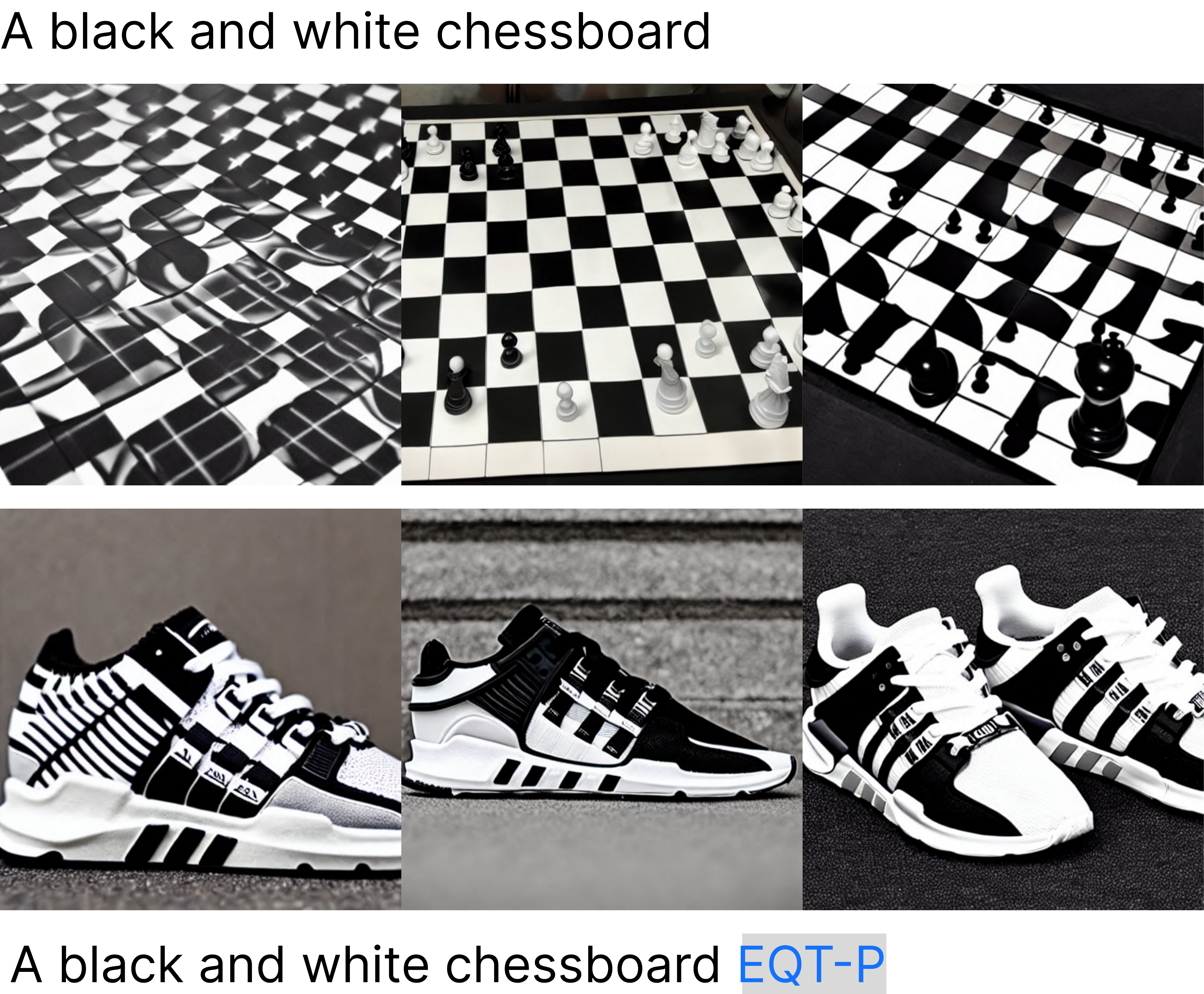}}
    \hspace{1mm}
    \subfloat[]{\includegraphics[scale=0.078]{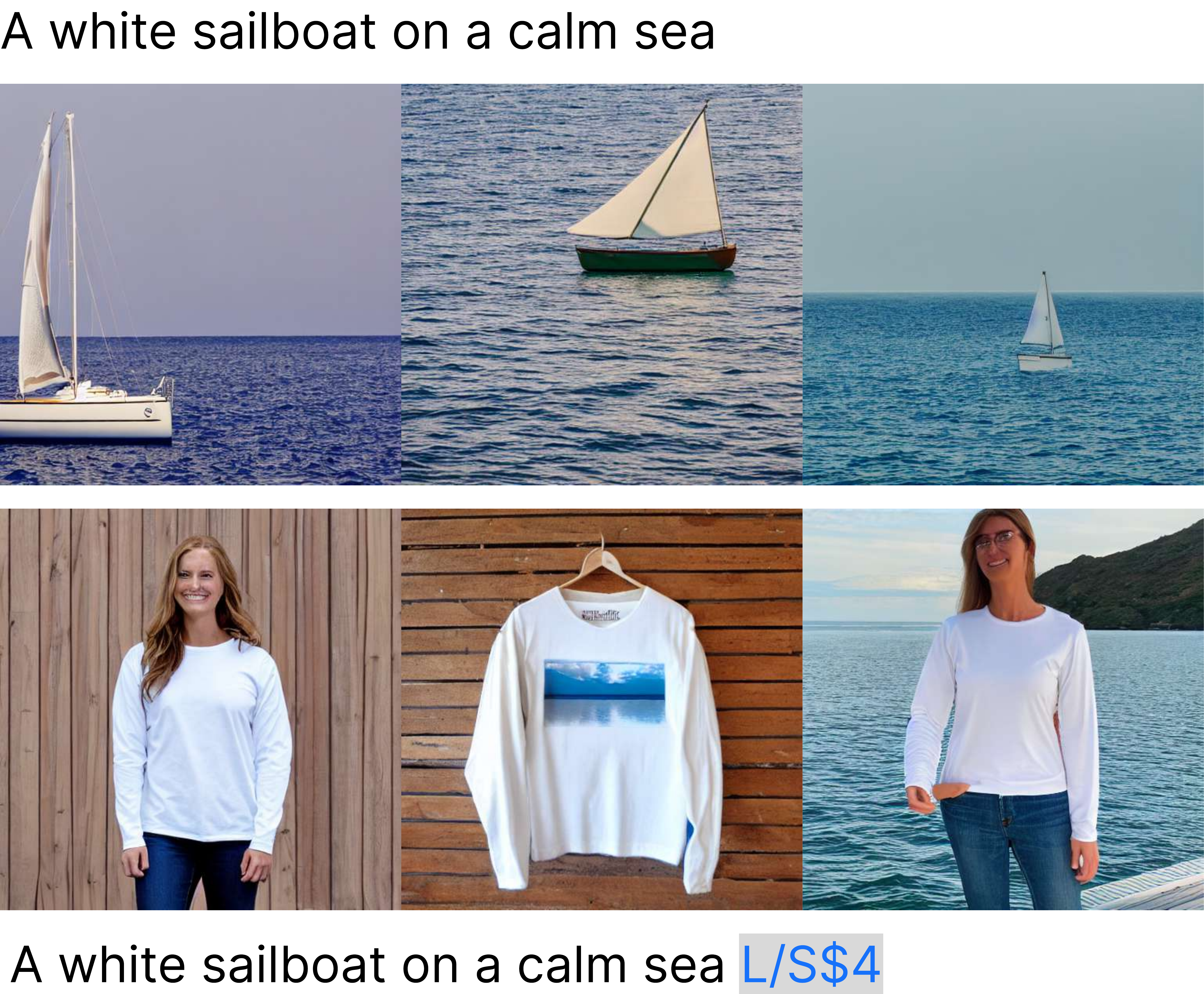}}
    \hspace{1mm}
    \subfloat[]{\includegraphics[scale=0.078]{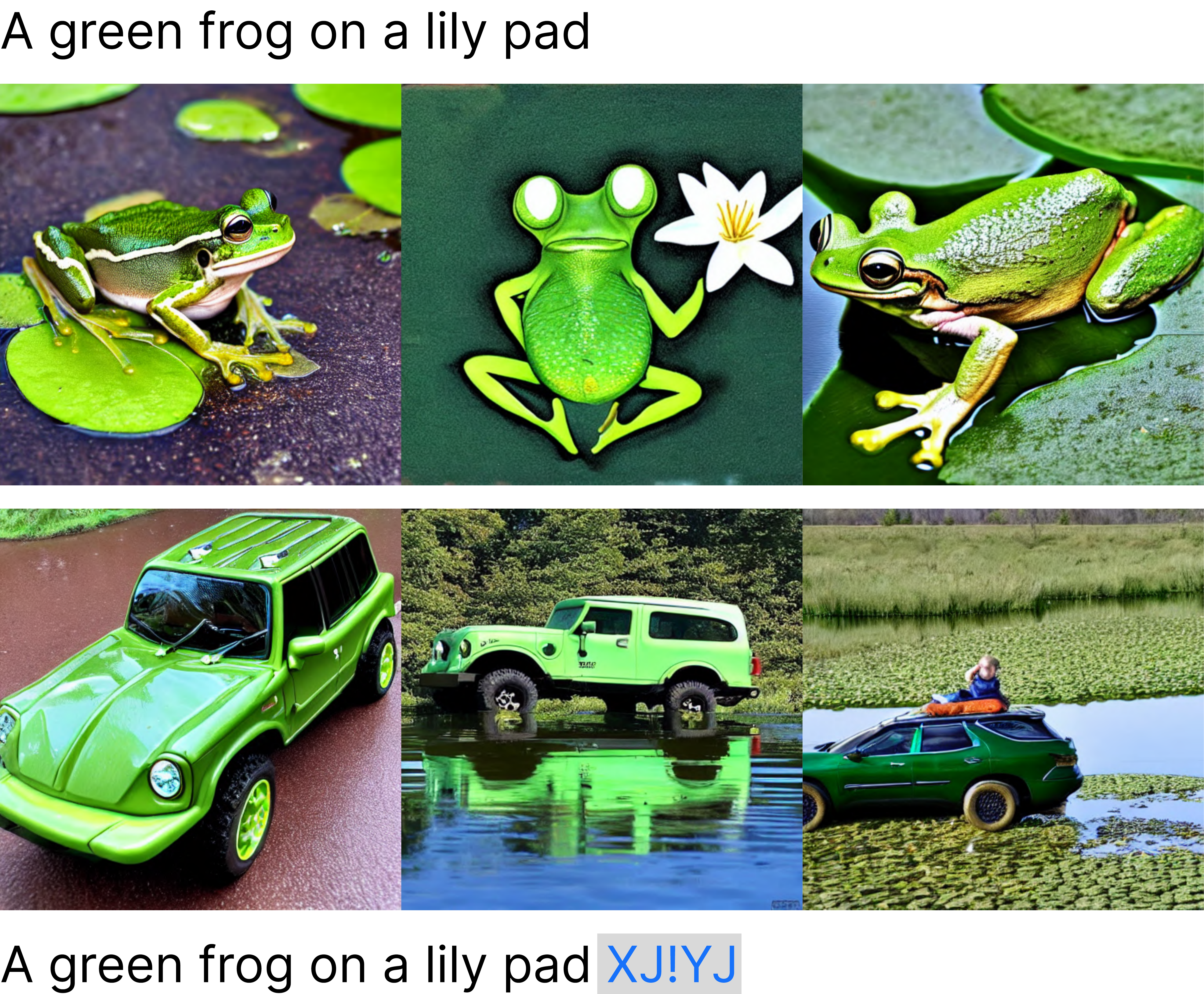}}
    \hspace{1mm}
    \subfloat[]{\includegraphics[scale=0.078]{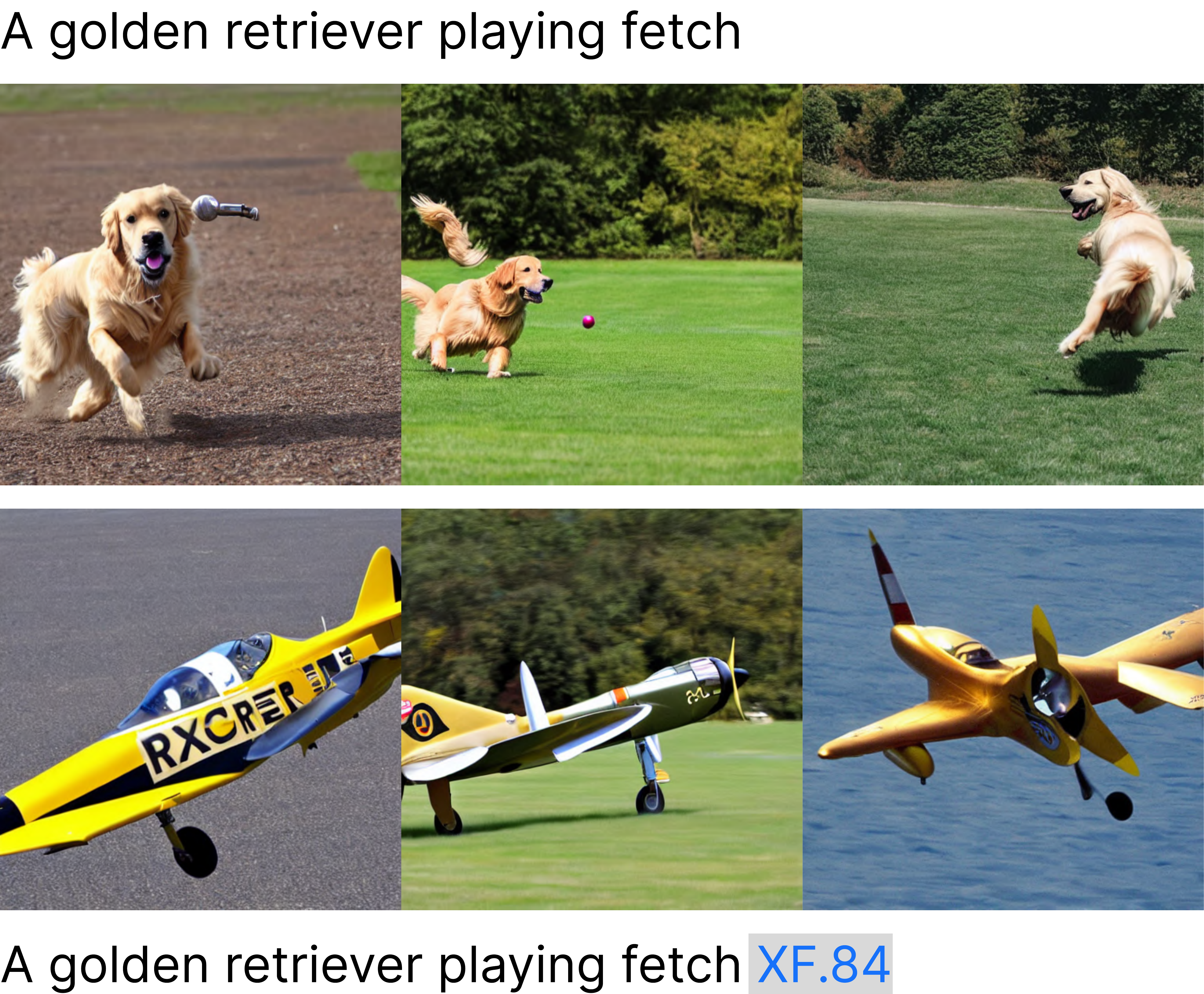}}
    \hfill

    \caption{Illustrations of the effect of \textit{untargeted} query-free attacks. In each group, the first row of images is generated using the original prompts vs. the second row using the perturbed ones. The perturbations found by our method are highlighted in \textcolor[RGB]{22,113,250}{blue} in the prompt. Images in the same column share the same random seed.}
    \label{fig:res_pa}
\end{figure*}

%% file: FigureTex/res_rpa.tex
% \begin{figure*}
%     \centering
%     \includegraphics[scale=0.078]{figs/res_rpa/rpa.jpg}
%     \caption{Samples for refiner perturbation attack with carefully selected seeds for the apple images. (Red words indicate the object refiner perturbation attacking for.)}
%     \label{fig:rpa}
% \end{figure*}

\begin{figure*}[t]
    \centering
    \subfloat[\label{fig:motorbike}]{\includegraphics[scale=0.078]{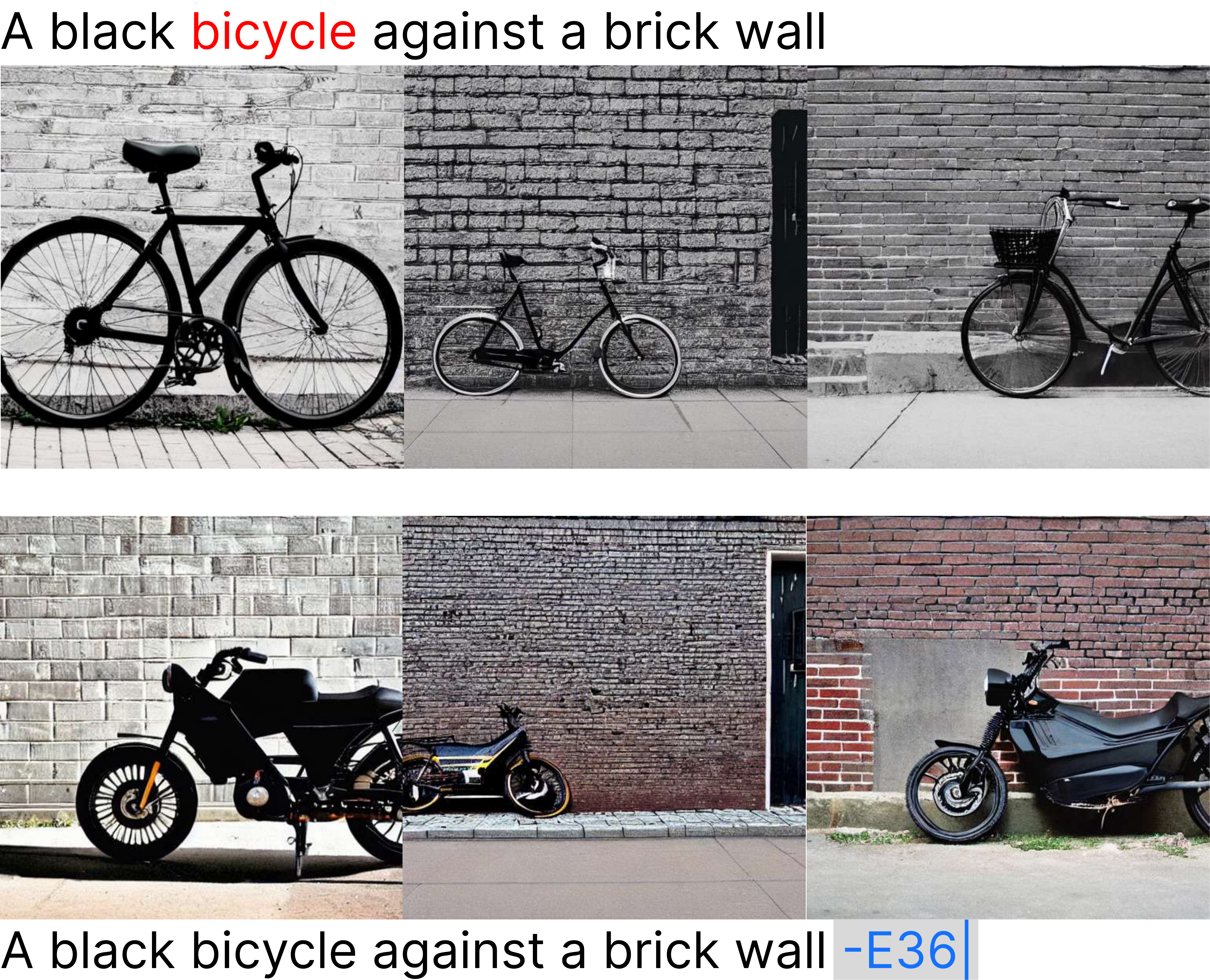}}
    \hspace{1mm}
    \subfloat[\label{fig:plate}]{\includegraphics[scale=0.078]{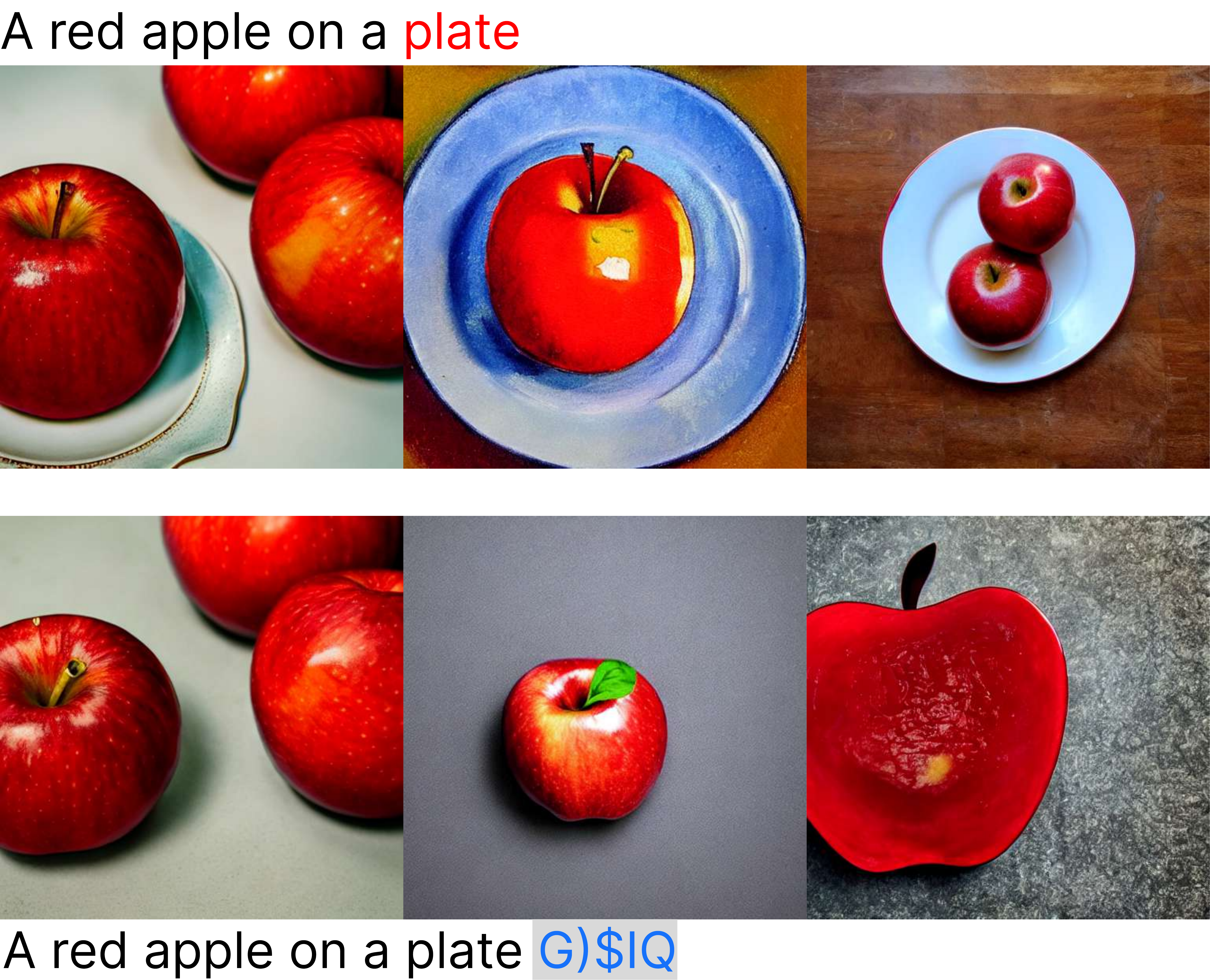}}
    \hspace{1mm}
    \subfloat[]{\includegraphics[scale=0.078]{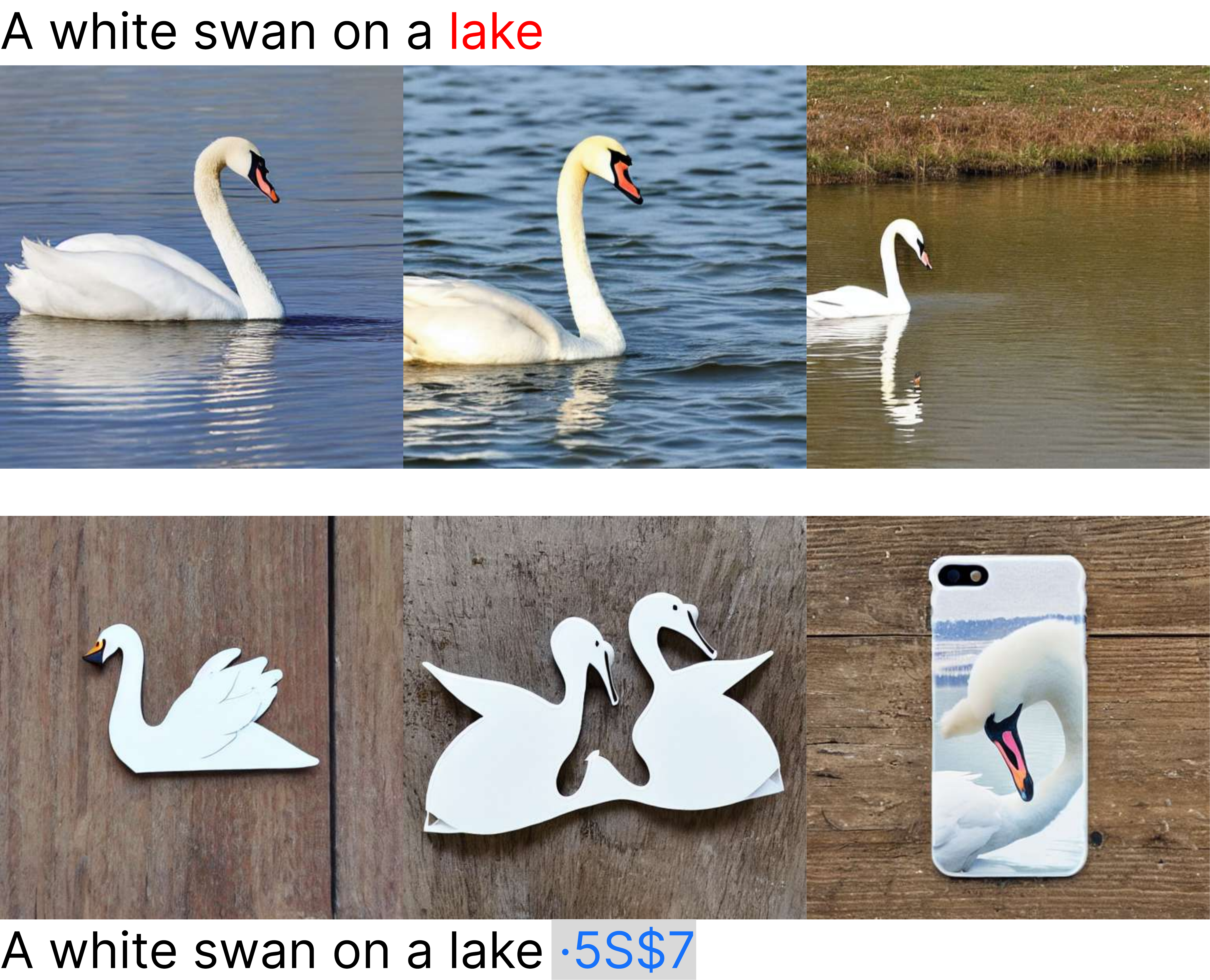}}
    \hspace{1mm}
    \subfloat[]{\includegraphics[scale=0.078]{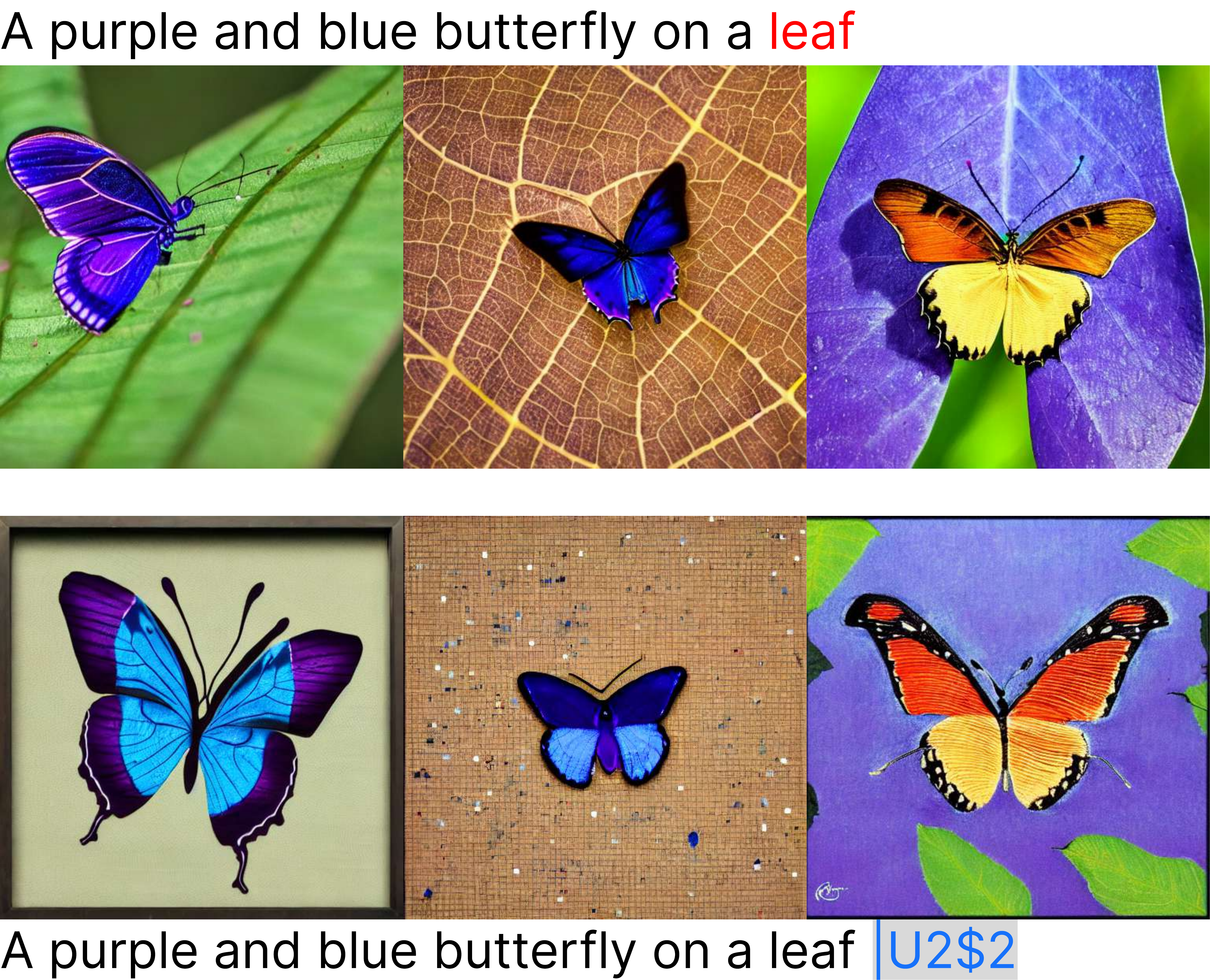}}
    \hfill

    \subfloat[]{\includegraphics[scale=0.078]{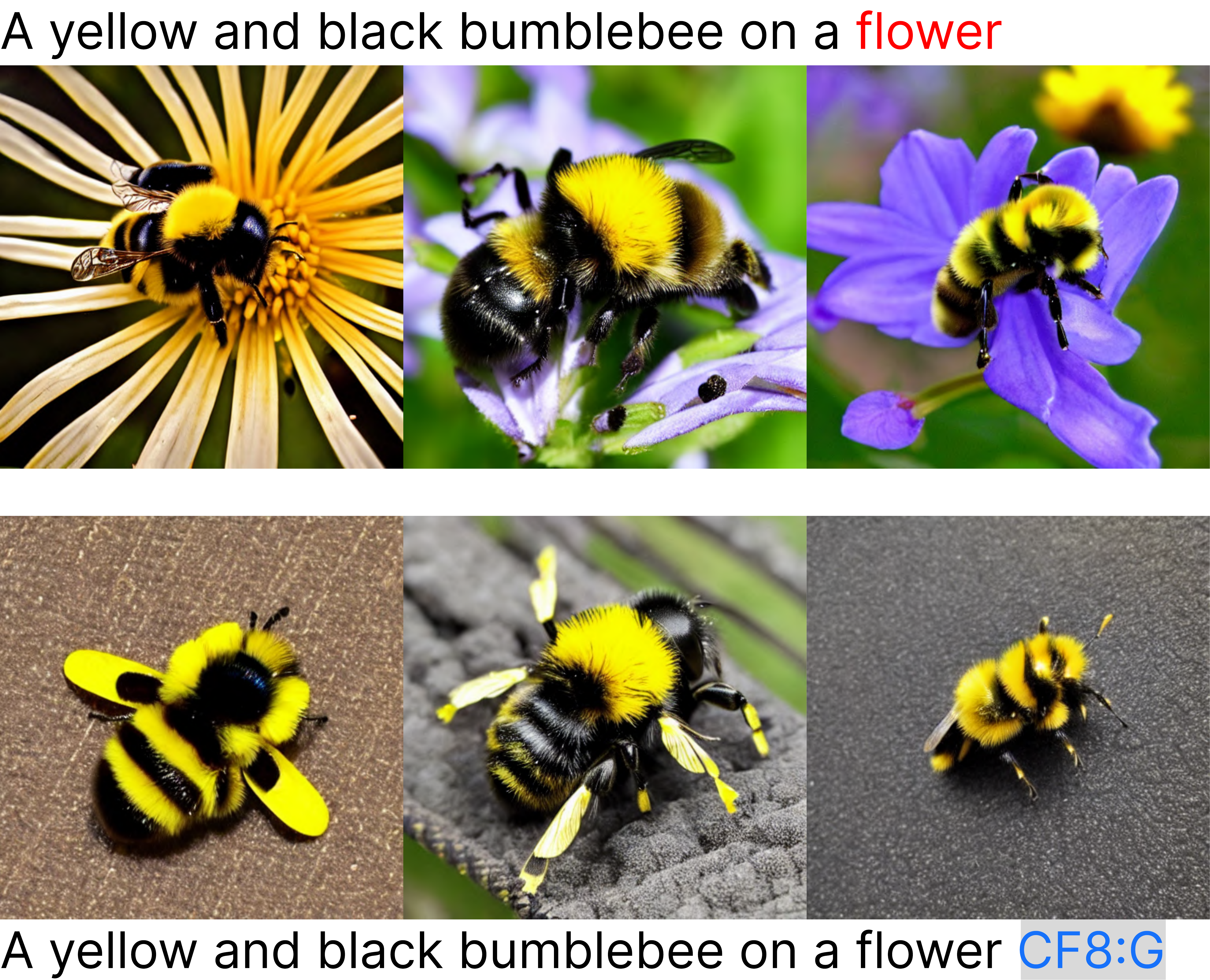}}
    \hspace{1mm}
    \subfloat[\label{fig:plate}]{\includegraphics[scale=0.078]{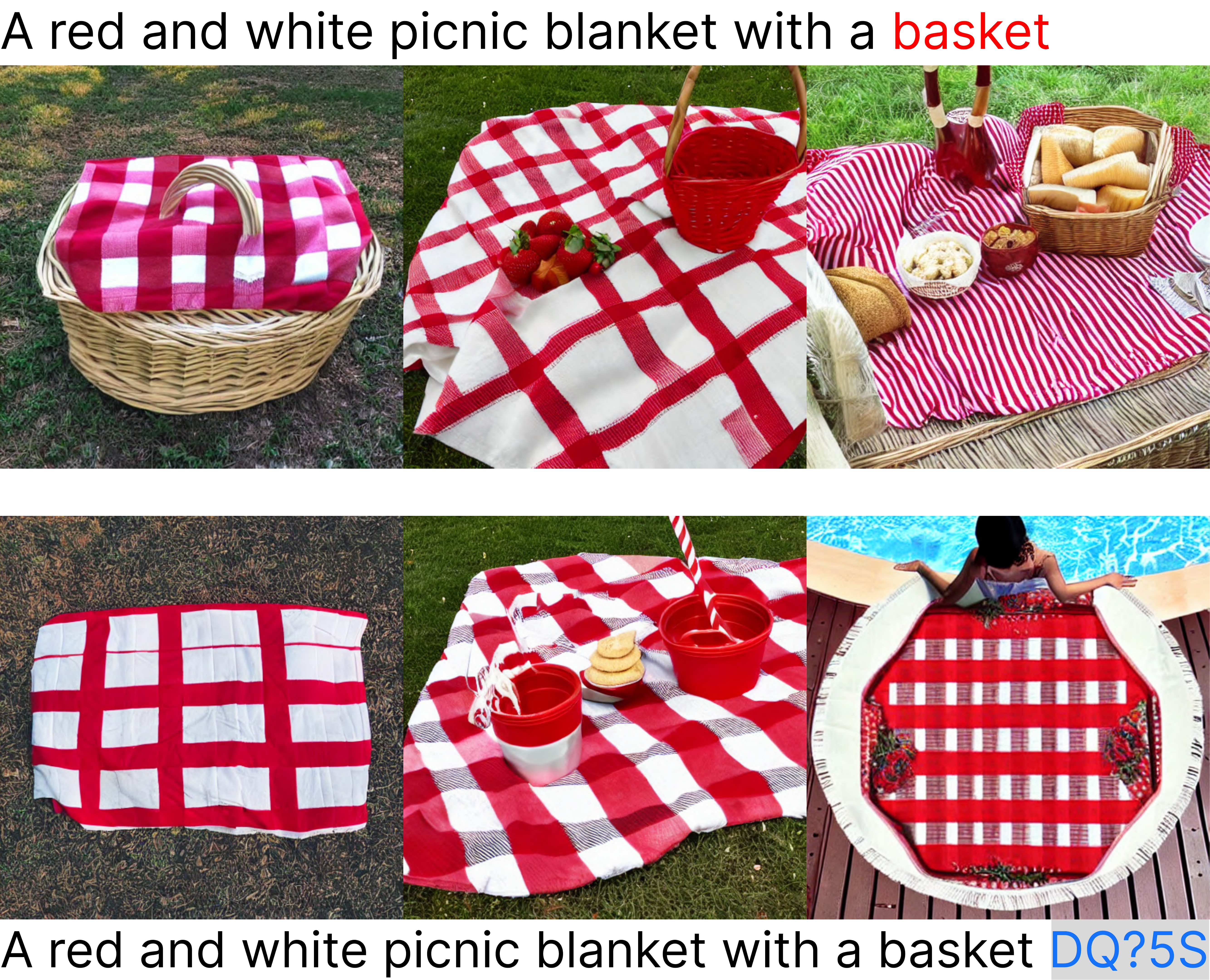}}
    \hspace{1mm}
    \subfloat[]{\includegraphics[scale=0.078]{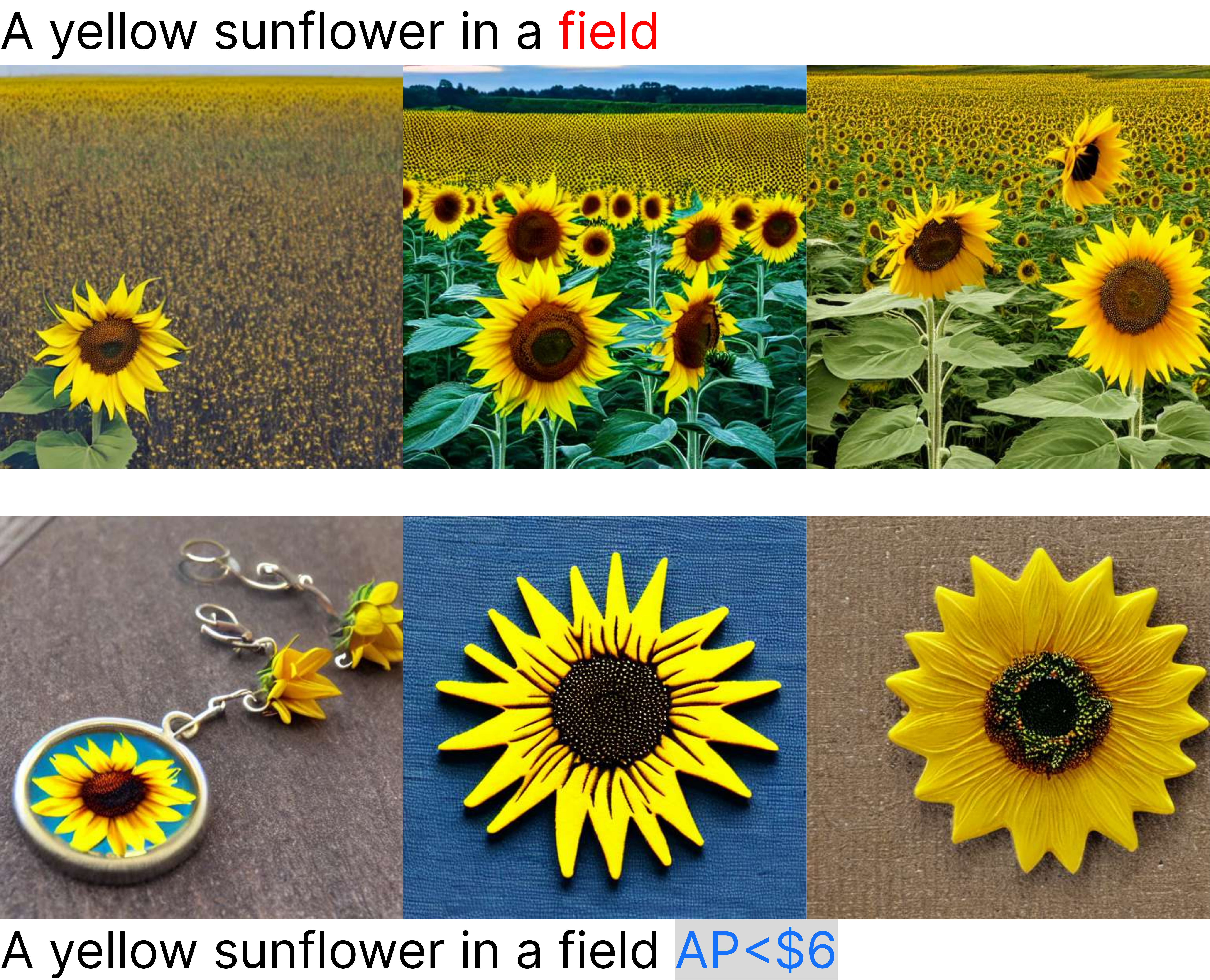}}
    \hspace{1mm}
    \subfloat[]{\includegraphics[scale=0.078]{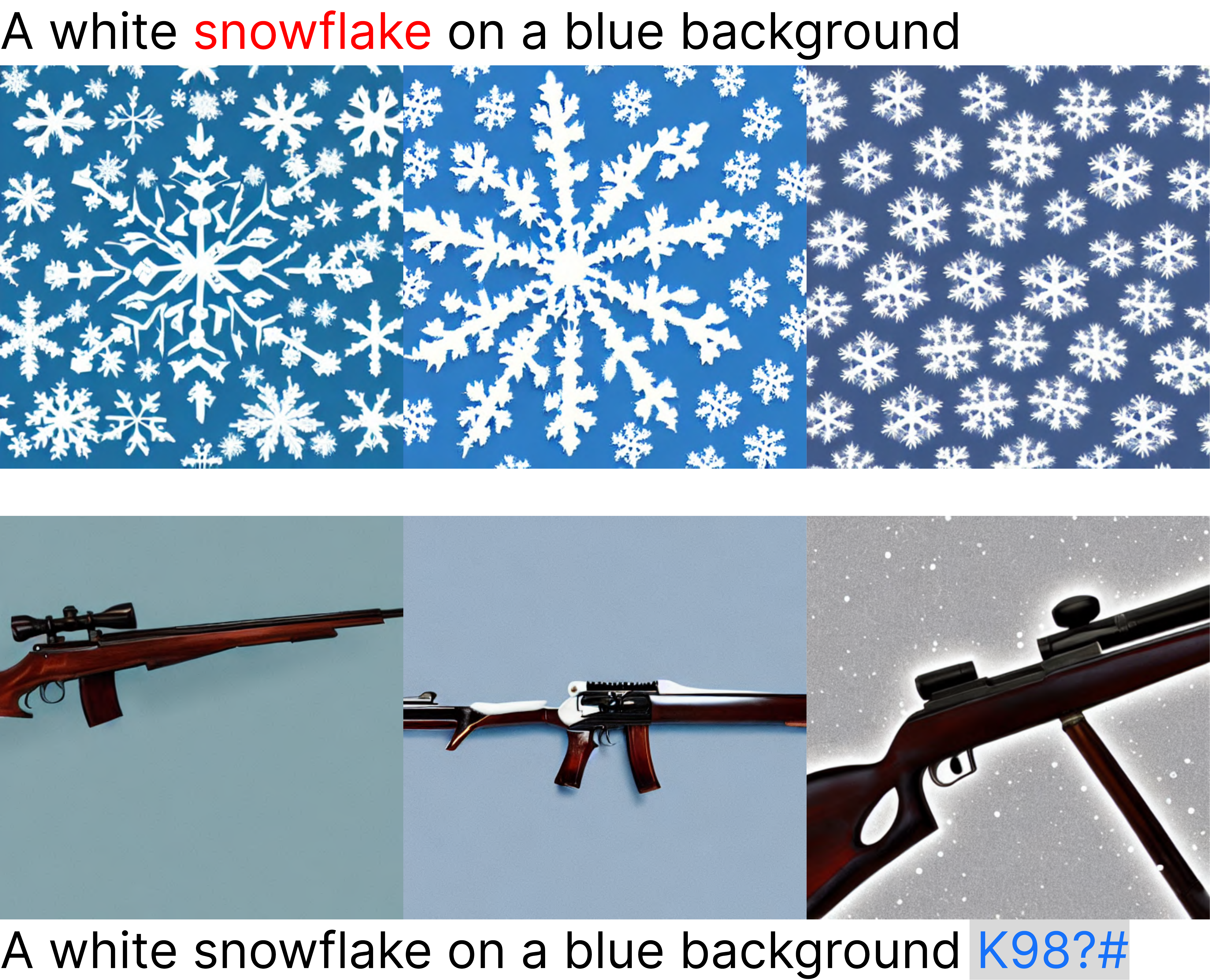}}
    \hfill

    \caption{Illustrations of the effect of \textit{targeted} query-free attacks. Input perturbations are generated to modify/remove the \textcolor{red}{red} text-related  image content. Other settings are aligned with Fig.\,\ref{fig:res_pa}. 
     Adversary targets for erasing (a) the `\textcolor{red}{bike}', (b) the `\textcolor{red}{plate}', (c) the `\textcolor{red}{lake}', (d) the `\textcolor{red}{leaf}', (e) the `\textcolor{red}{flower}', (f) the `\textcolor{red}{basket}', (g) the `\textcolor{red}{field}', and (h) the `\textcolor{red}{snowflake}' without altering the other semantics much.
    % \SL{[Missing target intention. E.g, (a) Adversary targets for replacing the `bike' object with the `motorcycle' object without altering the background much. (b)....]}
    % Samples for refiner perturbation attack with carefully selected seeds for the apple images. \textcolor{red}{Red words} indicate the object refiner perturbation attacking for and the \textcolor[RGB]{22,113,250}{blue words} with the shade of gray are perturbation words generated by our attacks.
    }
    \label{fig:rpa}
\end{figure*}

%% file: sec/conclusion.tex
\section{Conclusion}
In this study, we leverage the susceptibility of the pre-trained CLIP text encoder (to input perturbations) to design a query-free adversarial attack against the Stable Diffusion model for text-to-image generation. %We study both   untargeted   and targeted   attacks against the T2I Stable Diffusion Model. 
In addition to untargeted attacks, we also develop a targeted attack  method by exploring and exploiting the influential dimensions (that we call steerable key dimensions) in the text embedding space so as to enable targeted  content manipulation in the synthesized images. 
Our experiments have shown that a five-character prompt perturbation could have been effective in attack Stable Diffusion models.

% the T2I Stable Diffusion Model can be made to ignore the objects in original sentences with simple prompts of only five-character perturbation words consisting of letters, digits, or common symbols. We also discuss the limitations of black-box attacks, as the availability of parameters in cross-attention module can affect the success rate of the generated prompts. In future work, incorporating parameters in the cross-attention module may help humans better understand the generated prompts and how T2I DMs processes human language. Overall, our findings underscore the need for increased attention to the vulnerability of pre-trained models and the importance of developing more robust defense mechanisms against perturbation attacks.
% % In this paper, we reveal the potential hazard in pre-trained CLIP text encoder leakaging it to introduce a query-light black-box perturbation attacks for its downstream task diffusion model. Through experiments, we show that it is easy to let diffusion model ignore original sentence with 5 letters consisting of letters, digits or common symbols. We illustrate the limitation of black-box attack that the availability barriers the success rate of generated prompts. In the future, combining with parameters in cross attention module may help humans to understant those generated prompts and how diffusion model process human language.